\documentclass[10pt,twoside,a4paper]{article}
\usepackage{graphicx,color}
\usepackage[utf8]{inputenc}
\usepackage[T1]{fontenc}
\usepackage{bbm}
\usepackage{color}
\usepackage{amssymb}
\usepackage{multirow}
\usepackage{lineno}
\usepackage{hyperref}
\usepackage{algorithm}
\usepackage[noend]{algpseudocode}
\usepackage{amsmath}
\usepackage{subfig}
\usepackage{arydshln}
\usepackage{amsthm}
\usepackage{authblk}

\usepackage{listings}
\providecommand{\keywords}[1]{\textbf{\textit{Index terms---}} #1}

\begin{document}


\title{A redescription mining framework for post-hoc explaining and relating deep learning models}

\author[1,2]{Matej Mihelčić\footnote{Corresponding author: matmih@math.hr}}
\author[2,3]{Ivan Grubišić}
\author[2,3]{Miha Keber}
\affil[1]{Department of Mathematics, Faculty of Science, University of Zagreb, Bijenička cesta 30, 10000 Zagreb, Croatia}
\affil[2]{Ruđer Bošković Institute, Bijenička cesta 54, 10000 Zagreb, Croatia}
\affil[3]{Faculty of Electrical Engineering and Computing, University of Zagreb, Unska cesta 3, 10000 Zagreb, Croatia}


\maketitle

\begin{abstract}
 

Deep learning models (DLMs) achieve increasingly high performance both on structured and unstructured data. They significantly extended applicability of machine learning to various domains. Their success in making predictions, detecting patterns and generating new data made significant impact on science and industry. Despite these accomplishments, DLMs are difficult to explain because of their enormous size. In this work, we propose a novel framework for post-hoc explaining and relating DLMs using redescriptions. The framework allows cohort analysis of arbitrary DLMs by identifying statistically significant redescriptions of neuron activations. It allows coupling neurons to a set of target labels or sets of descriptive attributes, relating layers within a single DLM or associating different DLMs. The proposed framework is independent of the artificial neural network architecture and can work with more complex target labels (e.g. multi-label or multi-target scenario). Additionally, it can emulate both pedagogical and decompositional approach to rule extraction. The aforementioned properties of the proposed framework can increase explainability and interpretability of arbitrary DLMs by providing different information compared to existing explainable-AI approaches.

\end{abstract}



\section{Introduction}

Deep learning models \cite{DLSurvey} (DLMs) are inspired by complex interactions of neurons in a human brain and are capable of memorizing information, detecting patterns, predicting and summarizing information (potentially from heterogeneous sources), generating new data and much more. They have been applied to analyses of both structured and unstructured data. Learning on tabular data, time series prediction, image classification and segmentation, audio and video analyses are only a few applications.  This made them an important driving force of advancements in science and industry.

However, DLMs must often contain thousands of neurons grouped in different layers, to obtain satisfactory performance. This makes majority of DLMs de facto black boxes whose function is incredibly difficult to understand  \cite{EBM,BodriaGGNPR23}.

Current work that aims to increase understanding of black box DLMs creates understandable rule-based model that explains predictions made by deep neural networks (DNNs) e.g. \cite{DLRExtr,chen2020denas}, use general purpose tools to explain machine learning models, such as LIME \cite{LIME} or SHAP \cite{Shap}, e.g. \cite{kokalj-etal-2021-bert,rabold2019enriching} or propose novel models for explaining DNNs such as PatternNet and PatternAttribution \cite{Kindermans2018}. Perturbation and gradient-based methods, methods that utilize decomposition of predictions and trainable attention models can also aid in explaining DNNs predictions. Various approaches aim to allow analyses and visualization of neurons, groups of neurons and layers \cite{ExpAI}. 

In this work, we present a framework based on an unsupervised data mining technique called redescription mining \cite{RamakrishnanCart,GalbrunBW,MihelcicFramework,MihelcicRMRF}, designed to increase understanding of an arbitrary DLM and allow relating different deep learning models. The output of redescription mining is highly interpretable, providing entity descriptions and revealing complex attribute associations. For this reason, redescription mining is increasingly used in various domains \cite{RamakrishnanCart,MaladiesRM,MihelcicADNI, ReynaudDef, GalbrunLimitConditions, DevianceRM, MihelcicADMetals, ZoographicRegionsRM}. Recently, the task was also applied as an integral part of the methodology designed to increase understanding of convolutional networks applied to the problem of spatiotemporal land cover classification \cite{RmForDL,MergerNew}. Here, redescription mining was tasked to relate neuron activation levels of convolutional network with spatiotemporal patterns contained in the input data. Such relations could not be obtained without redescriptions, the result of redescription mining. 
Redescriptions are tuples of rules which are in equivalence relation. They have a rich query language, allowing the use of negation, conjunction and disjunction logical operators, which allows describing highly non-linear properties (e.g. the XOR function).


The framework proposed in this work is the first dedicated model-agnostic redescription mining based algorithm for interpretability and explainability of DLMs. It enables relating different layers of one DLM (intra-network), or relating layers of different DLMs (inter-network). Utilizing redecriptions, allows analysing predictions made by DLMs, discovering the role of individual neurons and analysing their associations, understanding the effects of utilizing different model parameters during training, detecting similarities between different DLMs or changes caused by performing transfer learning. The proposed framework uses relatively short descriptions with rich query language (using conjunction, disjunction and negation logical operators). It is naturally designed for parallel computation. This allows obtaining redescription-based descriptions, in reasonable time on small -- $30$ computing thread servers, even for selected layers of some state of the art DLMs, such as AlexNet \cite{Krizhevsky2012ImageNetCW}, EfficientNet \cite{EffNet} and BERT models \cite{BERT}. Utilizing more sophisticated machines would allow analysing even larger, more complex models. The proposed tool, aimed to be used by researchers, developers and experts studying DLMs and by researchers studying problems from various domains, such as biology, medicine, economy which increasingly use DLMs in their work, could foster many new scientific discoveries.

As it is demonstrated in Section \ref{sec:exp}, the tool can be used to understand a function of a neuron or a group of neurons in a DLM by relating the target group of neurons to the domain knowledge data (e.g. blood tests, cognitive ratings, genetic factors). Understanding of the function of the discovered subgroups of related neurons from several layers of one DLM or from different DLMs can be additionally increased through the analyses of the distribution of target class on the set of re-described entities.  In case of bad predictions, wrongly classified entities can be used to locate redescriptions that describe them. This reveals  neurons of DLMs that potentially require adjustments and domain properties that need to be further incorporated or augmented during DLM tuning, training and evolution. The procedure can also be used to study structural properties of DLMs. For example, detecting that one neuron from some layer of DLM can have a similar function as a set of neurons from the corresponding layer of some other DLM.  Information about structural properties of DLMs can be utilized in studying DLMs training process, transfer learning, effects of different random initializations on learning a target architecture, the process of distillation \cite{Distilation,FitNets} and many other fundamental properties of DLMs. 

Short summary of scientific techniques essential in the creation of the proposed approach: the redescription mining, Predictive Clustering trees and the CLUS-RM algorithm is provided in Section \ref{sec:prelim}. The related work to the proposed approach is described in Section \ref{sec:rel}. Methodology is thoroughly described in Section \ref{sec:methodology}, complexity and scalability analyses of the approach is provided in Section \ref{sec:complAndScal}. A set of experiments showing: a) the ability of the proposed approach to detect related properties of different DLMs, b) be used to explain predictions of DLMs using rules and redescriptions, c) can be used to explain and related layers of one or multiple DLMS, that it extends the functionality of existing approaches and outperform them, d) that the approach can create redescriptions containing very useful knowledge for analyses of various properties of DLMs, is provided in Section \ref{sec:exp}. Discussion analysing advantages and disadvantages of the proposed approach is given in Section \ref{sec:discuss} and the conclusions of the provided study are summarized in Section \ref{sec:conclusion}.

\section{Preliminaries}
\label{sec:prelim}

Redescription mining is an unsupervised data mining task that aims at discovering tuples of rule-based descriptions, where all rules in a tuple describe mutually similar or desirably the same subset of data entities. Such tuples, called redescriptions, are discovered for various subsets of input entities.  Each description in a tuple offers a point of view (perspective) that increases understanding of a targeted subset of entities. Redescriptions allow detecting regularities and associations between attributes (features) since rule-based descriptions in each tuple are in an equivalence relation. Redescription mining algorithms use a tabular dataset $\mathcal{D}$ containing a set of attributes $\mathcal{A}$, grouped into one or more views $W_i \in \mathcal{W},\ i\in \{1,2, \dots, |\mathcal{W}|\}$ describing a set of entities $E$. Views usually represent logical points of view on the available entities. Each view $W_i$ is a table of entities described by attributes $\{A_1, ..., A_{M_i}\} \subset \mathcal{A}$ of dimension $|E| \times M_i$. E.g. living organisms or patients can be characterized by their phenotypic, genomic or proteomic properties. 
Each attribute belongs only to one view. A redescription $R = (q_1, q_2, \dots, q_n)$ is a tuple of rules (also called queries) $q_1,\dots q_n$. Each rule $q_i$ contains only attributes from the corresponding view $W_i$. A redescription set (RS) is denoted $\mathcal{R}$.

A set of entities described by a query $q_i$ is denoted  $supp(q_i)$, whereas a set of entities described by a redescription $supp(R) = \cap_{i=1}^{n} supp(q_i)$. Redescription accuracy is quantified with the Jaccard index. 
\begin{equation}
\label{measures:jacc}
J(R) = \frac{|\cap_{i=1}^{n} supp(q_i)|}{|\cup_{i=1}^n supp(q_i)|} 
\end{equation}

Jaccard index of $1.0$ denotes that each query forming a redescription describes only entities from a redescription support set and no other entities from the dataset. Thus, $q_1$ is true if and only if $q_2$ is true etc. Jaccard index value lower than $1.0$ denotes that the equivalence relation is not perfect. There must exist exceptions, entities described by some queries and not described by at least one query forming some redescription. Statistical significance of a redescription is quantified by a $p$-value and computed from a Binomial or Hypergeometric distribution. We utilize common formula derived from Binomial distribution.
\begin{equation}
\label{measures:pval}
p(R)=\sum_{k=|supp(R)|}^{|E|} {|E|\choose k}(\prod_{i=1}^{n} p_i)^k\cdot(1-\prod_{i=1}^{n} p_i)^{|E|-k} 
\end{equation}
\noindent $|E|$ denotes the number of entities in the dataset, $p_1 = |supp(q_1)|/|E|$, $p_2 = |supp(q_2)|/|E|, \dots, p_n = |supp(q_n)|/|E|$ are the marginal probabilities of obtaining $q_1$, $q_2, \dots, q_n$. More detailed description can be seen in the work \cite{GalbrunBW}. $attrs(R)$ denotes a set of all attributes contained in redescription queries. Machine learning model of type $\mathcal{C}$ with parameters $\mathcal{P}$ is denoted as $\mathcal{M}_{\mathcal{C}}^{\mathcal{P}}$. We use $L_k^{\mathcal{M}_{\mathcal{C}}^{\mathcal{P}}}$ to denote the $k$-th layer of the (deep) neural network $\mathcal{M}_{\mathcal{C}}^{\mathcal{P}}$ and $n_{i,k}\in L_k^{\mathcal{M}_{\mathcal{C}}^{\mathcal{P}}}$ to denote the $i$-th neuron in the $k$-th layer of the (deep) neural network $\mathcal{M}_{\mathcal{C}}^{\mathcal{P}}$.

Redescription $R_{ex} = (q_{1,ex}, q_{2,ex})$, obtained with the proposed methodology: \par \noindent
$q_{1,ex}: 0.98\leq n_{23,3}\leq 1.41\ \wedge\ 1.14\leq n_{8,3}\leq 2.32\ \wedge\  0.74\leq n_{21,3}\leq 1.61 $\par \noindent 
$q_{2,ex}: 1.0\leq \text{EcogOrgan}\leq 2.5\ \wedge\ 23.0\leq \text{RAVLT\_IMMEDIATE}\leq 55.0\ \wedge\ 15.0\leq \text{ADAS13}\leq 23.0\ \wedge\ 4.84\leq \texttt{FDG}\leq 6.43\ \wedge\ 2 \leq \text{FAQ}\leq 9$, relates neuron activations from the penultimate (third) layer of the feed forward neural network, trained to predict dementia levels of suspected patients, with the domain knowledge containing various cognitive tests and biological indicators taken to assess the cognitive dementia and the Alzheimer's disease of a set of suspected patients. The crux is that the obtained redescription exposes an equivalence relation between the discovered queries, thus the neuronal activation of the described group of neurons co-occurs if and only if cognitive tests and biological indicators have their values in the predetermined intervals. The strength of this equivalence is measured by the Jaccard index, which is $0.75$ for the example redescription. Thus, both queries are true for $3/4$ of all patients for which either of the above queries is true. Values of the cognitive tests and indicators, combined with the Jaccard index information, showcase that the described group of neurons, with the predetermined activations, jointly mostly ($3/4$ of the times) describe cognitively impaired patients. Redescription describes $12$ patients in total, and for $16$ patients at least one of the aforementioned queries are true. Example redescription is statistically significant with $p = 7.85\cdot 10^{-13}$. This demonstrates that the example redescription could not be easily obtained by random pairing of queries describing similar number of entities.  Functions of the discovered set of neurons can be definitely confirmed by analysing the distribution of target labels of all patients described by either query. This distribution contains a patient diagnosed with the Alzheimer's disease, seven patients diagnosed with late cognitive impairment and eight patients diagnosed with early cognitive impairment. Thus, no cognitively normal people or people with small memory concern are contained in the group. Using this fact, and the value intervals of cognitive tests and biological indicators, we can deduce that the combination of discovered neurons with the predefined activations are predominantly involved in the detection of patients with early and late cognitive impairment. 

Predictive Clustering trees \cite{PCT,Kocev} are a generalization of decision trees, where a variance reduction heuristic is used to choose the split point and the prototype function is used to compute the centroid of each set of entities belonging to the node of a tree (node cluster). Centroids are used to make predictions and allow computing distances between different clusters. Utilization of centroids in the heuristic function used to compute the splitting point, allows for the use of both data attributes and target labels to define clusters (sets of entities belonging to a node in a tree). It also allows using various types of target labels (classes, targets, multi-label, multi-target or hierarchical). Predictive Clustering trees are implemented in the CLUS framework \cite{CLUS}.
The CLUS-RM \cite{MihelcicRMRF} is the redescription mining algorithm that uses two data tables with disjoint sets of attributes (data views) to produce redescriptions. It utilizes Predictive Clustering trees to obtain rules used to construct redescriptions. Predictive Clustering trees are trained in CLUS-RM alternations, where nodes of one PCT, trained on one data table, are used as targets to train the next PCT using the second data table. The process utilizes multi-target regression and multi-label classification abilities of PCTs. In each iteration, a pair of PCTs and potentially a supplementing random forest of PCTs are used to create new redescriptions.  The algorithm uses a redescription set optimization procedure to construct the final set of redescriptions that is presented to the user. The CLUS-RMMW \cite{MWRM} is the generalization of the CLUS-RM algorithm that enables creating redescriptions on datasets containing $k$ views. The algorithm first computes a set of incomplete $2$-view redescriptions, utilizing available views in a pairwise fashion, and then completes these redescriptions by iteratively training PCTs on each remaining view. The approach internally stores a  candidate set of redescriptions, which is iteratively improved, updated and from which a final set of output redescriptions is created using a redescription set optimization procedure. The candidate set is mostly significantly larger than the output set, and larger candidate sets mostly allow obtaining output redescription sets of higher quality. Codes of the CLUS-RM and CLUS-RMMW approaches are available in \cite{CLUSRMSoft}.

\section{Related work}
\label{sec:rel}
The approach proposed in this manuscript aims to explain, interpret DLMs and allow relating various DLMs using redescriptions -- tuples of rules. It is related to approaches that aim to explain DNNs using rule extraction \cite{DLRExtr,SurvRule}. These models either build rule sets that aim to explain predictions of neural networks or DNNs, e.g. \cite{RulesTrANN}, describe neurons or layers by utilizing descriptions of neurons in previous layers, e.g. \cite{zilke2016deepred}, or try to determine what caused a given output of a neural network, e.g. \cite{ReverseEng}. Divide and conquer approaches \cite{DLRExtr,zilke2016deepred} are the first that have been applied to explain the actual DNNs, previous approaches were mostly used to explain simple artificial neural networks. Ribeiro et al. \cite{Anchors} created a model agnostic system to explain DLMs using rules of high precision called \emph{anchors}. Wang \cite{pmlrv97wang19a} developed a model agnostic rule-based approach called \emph{Hybrid Rule Sets}. It discovers subsets of data where rule-based model is closely accurate as the targeted black-box model. 

As opposed to aforementioned approaches that use rules to explain DLMs, the proposed approach uses redescriptions. The main advantage of using redescriptions over rules is largely increased generality of the approach. Since redescriptions are tuples of rules, the approach produces and is ultimately able to utilize both. In addition to being able to create a surrogate model that explains decisions made by some neural network or DNN, the proposed approach enables relating neurons and groups of neurons within one or between different DNN models. Moreover, the approach enables explaining neurons, groups of neurons and layers using original or supplementary data (target labels, knowledge bases etc.). Due to the fact that redescriptions consist of rules that are logically in an equivalence relation, the approach allows detecting domain properties that occur if and only if a particular neuron (or a group of neurons) fire with a predetermined intensity. The proposed approach does not depend on the architecture of a DLM and is applicable to fairly large DLMs, which are weak points of the majority of available rule-based approaches. 

Methods such as \texttt{SVCCA} \cite{SVCCA} and \texttt{CKA} \cite{CKA} allow computing similarity of different representations, however they do not produce interpretable rule-based descriptions that would explain the source of their similarity. The proposed approach closes this gap by detecting subsets of neurons in different DLMs with similar functions and  discovers associations between their neuron activations.

As authors are aware, M\'{e}ger et al. \cite{RmForDL, MergerNew} are the only works that attempted using redescription mining to explain convolutional networks applied to the problem of land cover classification.  The proposed tool is architecture specific and uses redescription mining as the out of the box tool, the latter hinders performance and reduces the number of neurons that can be described (as shown by the experiments performed in this work). Furthermore, the approach is limited to using only two views, neuron activations and original data, which precludes relating different architectures or using knowledge bases to enhance understanding of the targeted model.

Multi-view redescription mining \cite{MWRM} allows relating two or more models, however its use as an out-of-the-box tool enables describing only a limited number of neurons, mostly discovering their interactions, as shown by the performed experiments. Thus, the authors find it inadequate for explaining and relating complex DLMs. 


\section{Methodology}
\label{sec:methodology}

We present a methodology \texttt{ExItNeRdoM} (\emph{Explaining and interpreting (deep) neural models using redescription mining}). The presented post-hoc explainability methodology \cite{ExpSurv}, based on redescription mining, is specifically tuned to explain and relate neurons and groups of neurons of DLMs. It enables detecting functionally similar parts in different layers of the same or across different DLMs and detecting correspondence between levels of neuron activations and domain data attributes or target label values.
Using redescriptions instead of rules greatly increases the generality of the proposed approach since the methodology produces and can utilize both rules and redescriptions.  The approach allows providing both local and global type of explanation, explanation by example, cohort explanation, and provides text-based explanations in a form of redescriptions. It allows deep insight into the activations of groups of neurons and their relation to the target labels of the analyzed task. The methodology also allows detecting and analyzing wrongly assigned examples, which could be used to improve the model by focused training. Redescriptions, obtained by the approach, can be used as features to improve predictive performance of interpretable models \cite{MihelcicFeatures}, e.g. decision trees. 

Technical novelties of the proposed approach compared to related redescription mining approaches are: 
\par \noindent
a) Exhaustive descriptions of neurons individually and in interactions, logically grouped by the neuron $id$, allowing easy detection and exploration of knowledge of interest. Regular redescription mining approaches lack the logical division and have significantly lower description coverage of neurons, especially individually. 
\par \noindent b) Modified binning procedure that takes into account specificities of neuronal activations (low or no activations vs high activations) and allows obtaining redescriptions explaining neurons individually. The attribute binning in general form is only performed by the ReReMi approach \cite{GalbrunBW}, however single attributes are only used as a starting point for the greedy extension process, where more complex and accurate redescriptions are singled out. It is possible to limit the number of attributes in a query to $1$ with the following limitations: 1) no knowledge about the fact that the attributes represent neuronal activations, 2) no logical groupings of produced redescriptions, 3) no possibility to use more than $2$ data views, 4) obtaining redescriptions involving interactions of neurons requires starting the redescription mining anew with different settings.
\par\noindent  c) Use of rules describing individual particular neuron as targets, guiding the search towards redescriptions revealing interactions of a chosen neuron with other neurons in one or more layers of a DLM. This is the novelty compared to the workings of the redescription mining algorithm CLUS-RM \cite{MihelcicRMRF}, which does not have rules describing individual neurons readily available. The CLUS-RM algorithm \cite{MihelcicRMRF} incorporates attribute diversity as one of the criteria in the optimization process, however it does not guarantee that majority or all attributes will be represented. 
\par\noindent  d) Adaptation of the constraint-based methodology \cite{MihelcicADNI} to multi-view setting \cite{MWRM} and its incorporation into the proposed methodology, allowing selective search for neuronal interactions involving selected neuron from a DLM. Besides being restricted to two views, methodology \cite{MihelcicADNI} provides no logical grouping of the produced redescriptions using different constraint sets (or the process needs to be restarted for every constraint set). 
\par\noindent e) Incorporation of parallel (multi-threaded) computing, which allows performing explanations of multiple neurons (individually and in interaction) in different working threads, allowing the proposed methodology to produce significantly more detailed descriptions of studied DLMs, using competitive execution times.

\subsection{General overview}

\begin{figure}[ht!]
\includegraphics[width=\textwidth]{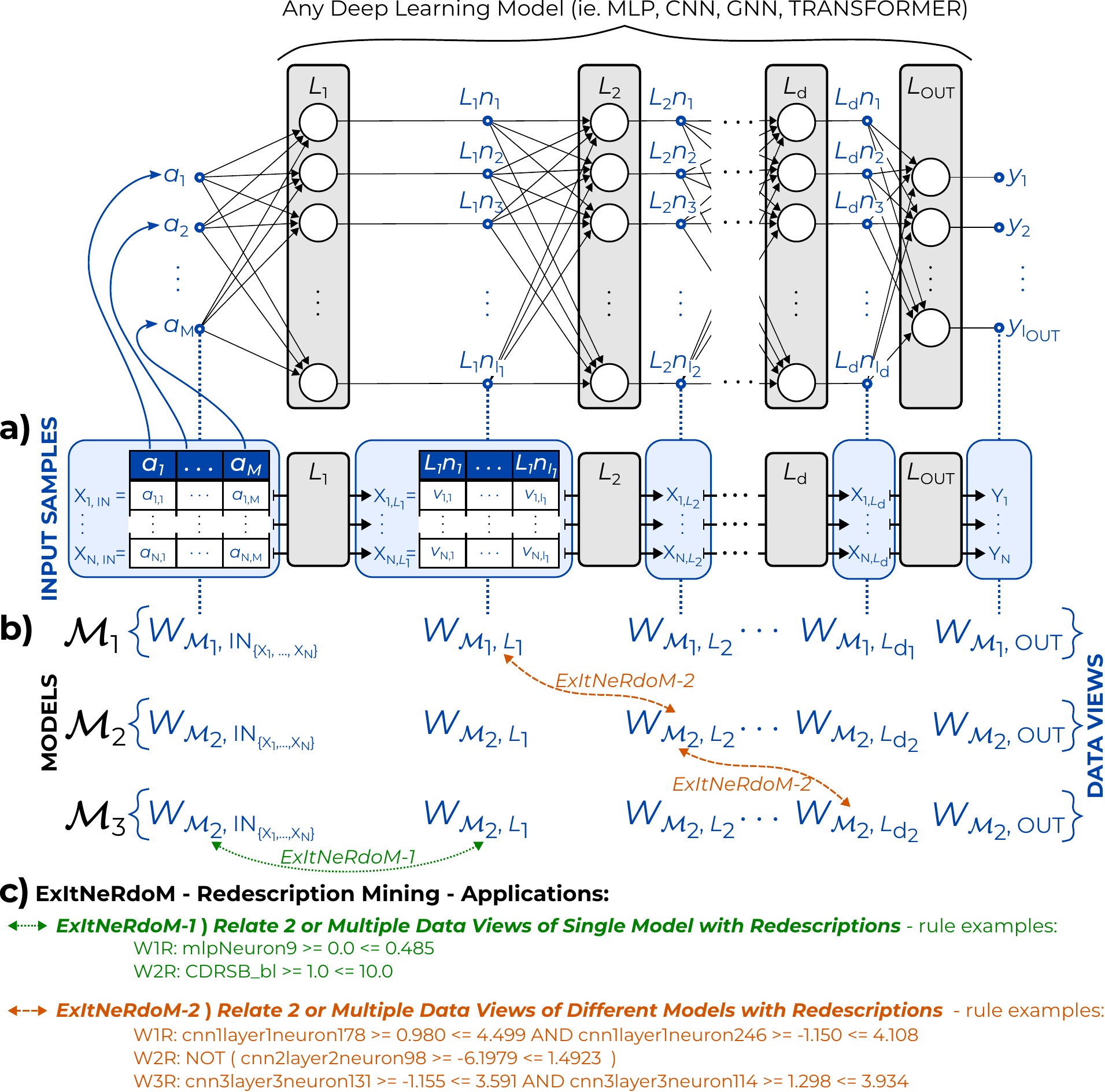}
\caption{ The overall computation flow of analysing DLMs. \textbf{a}) Main data input to the RM algorithm are views $W$ with cardinality of the entities $|E|=N$ as rows $\{X_1, … , X_{N}\}$ and columns as attributes $\{A_1, \dots, A_{M}\} \subset \mathcal{A}$. Attributes can be domain data features or hidden representation values, i.e. neuron activations after forward propagation of an entity through a trained DLM. Part \textbf{b}) Arbitrary views can be chosen as input to the \texttt{ExItNeRdoM}, even between different DLMs that in any way represent the same set of entities. $L_kn_i$ attribute represents $i$-th neuron activation in the $k$-th layer, i.e. $n_{i,k}$. \textbf{c}) are the outputs of the framework. Rule \texttt{ExItNeRdoM}-1 describes that MLP neuron $n_{in,9}$ activation is in $[0,0.485]$ \textbf{if and only if} the Cognitive Dementia Rating Sum of boxes, CDRS\_bl is in $[1,10]$. To understand the strength of the discovered equivalence,  we need information about $supp(\texttt{ExItNeRdoM}\text{-}1)$, and the corresponding Jaccard measure, see Section \ref{sec:exp}.}
 \label{fig:overview}
\end{figure}

Fig. \ref{fig:overview} contains the overview of the process of transforming the input DLM (e.g. MLP, CNN, GNN, Transformer) into the input data for the \texttt{ExItNeRdoM}, the general process of relating various layers of different DLMs and the examples of obtainable results. The methodology, optionally, takes the domain data, depicted as the first input view ($W_{\mathcal{M},IN}$). $a_1, \dots, a_M$ denote the domain data attributes (e.g. blood markers, pixels etc.). Zero or more domain input tables can be used to improve the understanding of input DLMs, where a subset can be used to train the DLMs and a part can be additional domain-level knowledge. Entities of interest (images, patients, etc.) are forwarded through the input DLMs, and the neuron activation weights in each layer ($n_{i,k}$) are recorded as new attributes for these entities (see columns $L_kn_i$, in part a) of Fig. \ref{fig:overview}, in Tables corresponding to views $W_{\mathcal{M}_{1,IN}},\dots \mathcal{M}_{M_1,OUT}$ from part b) of the same figure). All views use the same set of entities. Target labels can contain one or more multi-class or regression labels as attributes for each input entity (denoted in matrix form $\mathbf{Y_1, \dots, Y_N}$, see Fig. \ref{fig:overview} part a). The \texttt{ExItNeRdoM} approach can either take a sequence of DLM layer tables as inputs with zero or more domain data tables ($\emptyset$ denotes zero tables), or at least one domain data table and one target label table obtained by the DLM to mimic the pedagogical rule extraction. Arrows indicate that any combination of these views can be chosen as inputs to the \texttt{ExItNeRdoM} methodology. It is recommended to choose layers of DLMs that allow explaining the behaviour under investigation. E.g., the penultimate layer to understand the final predictions, intermediate layers to understand the impact of different architectures or initialization on weight re-distribution and organization etc. The approach uses the constructed views, adapting techniques from redescription mining (see Algorithm \ref{alg:ExpInt}) to describe entities with redescriptions, which provide information about associations between neurons of interest in a studied set of DLMs. 

The output redescriptions, obtained by the \texttt{ExItNeRdoM} methodology and presented in Fig. \ref{fig:overview}, part c), demonstrate interpretable, textual information about the input models provided by the approach. The \emph{ExItNeRdoM-1} from Fig. \ref{fig:overview}  relates the penultimate view of the MLP trained on the ADNI dataset and the view of the original attributes of these data. This redescription indicates that the MLP neuron $9$ will be within interval $[0, 0.4853]$ for $125$ original data entities if and only if the \emph{CDR sum of boxes} scores of these patients is in the interval $[1,10]$. This occurs with accuracy of $0.828$. The \emph{CDR sum of boxes} score in the interval of $[1, 10]$ occurs in patients with suspicion of some level of cognitive impairment: MCI, LMCI, AD \cite{o2008staging}, which gives immediate information to the domain expert. The domain expert will immediately notice there is a very small number of exceptions from this redescription and could study the distribution of neuron $9$ activations for the obtained group of entities, the distribution of target labels, the distribution of neuron $9$ activations on healthy control subjects or analyse the interactions with other neurons to deepen the understanding of the role of neuron $9$ in the process of diagnosing the Alzheimer’s disease. The \emph{ExItNeRdoM-2} is more complicated, given its task to relate the penultimate layers of $3$ different CNN architectures trained on the MNIST dataset. Here, one can observe that the given neurons with their activations activate in all $3$ networks for $784$ entities with accuracy of $0.612$, meaning there are around $304$ entities for which at least one rule is not valid. This redescription allows studying the redistributions of weights that occurred due to differences in initialization and architecture design. One can study post-hoc what pixels of images cause joined activations of neurons in all $3$ networks to study their similarities and differences, ultimately revealing the process of redistribution that occurred either due to the differences in architecture or in model initialization.

\begin{figure}[ht!]
\begin{center}
\includegraphics[width=0.8\textwidth]{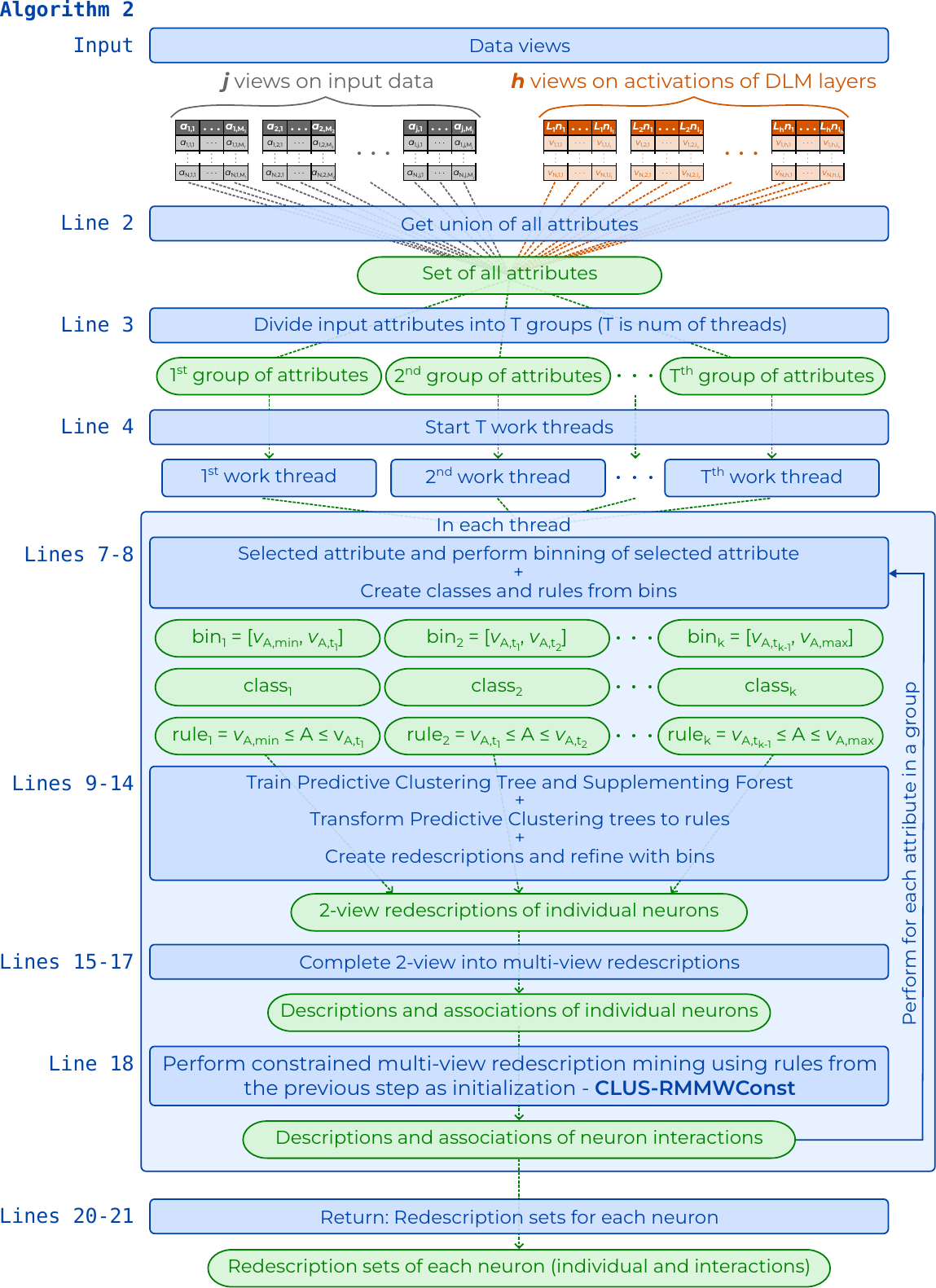}
\end{center}
\caption{A diagram of the \texttt{ExItNeRdoM} methodology.  Methodology can accept multiple views describing the same set of entities. Blue boxes are the main steps of the methodology, and the green boxes represent corresponding intermediate and final output results of the methodology. Lines on the left correspond to lines in Algorithm \ref{alg:ExpInt}. 
}
 \label{fig:diagram}
\end{figure}

\noindent The high level overview of the  \texttt{ExItNeRdoM} methodology can be seen in Fig. \ref{fig:diagram}. The \texttt{ExItNeRdoM} methodology can perform binning of all input attributes or just views containing neurons of interest. Rules obtained from the bins are related to rules describing other attributes and neuron representations using a sequence of steps involving training of Predictive Clustering trees (PCTs), joining rules from different views into redescriptions and performing the constraint-based multi-view redescription mining. Technical modifications required to perform each of these steps are described in Section \ref{sec:endetails}.
The methodology outputs two sets of redescriptions per attribute (neuron), one describing interactions of a targeted neuron with various activation intervals to neurons contained in other data views, and one containing description of interactions of the selected attribute (neuron) with attributes (neurons) contained in the same and other data views. The presented outputs can be used to obtain various information about the analyzed DLMs, this is further discussed in Sections \ref{sec:exp} and \ref{sec:discuss}.

\subsection{Technical details of the \texttt{ExItNeRdoM} methodology}

\noindent In this section, we present the technical details of the \texttt{ExItNeRdoM} methodology. Section \ref{sec:endetails} describes the building blocks of the methodology and explains their differences to the building blocks of the related multi-view redescription mining algorithm, the CLUS-RMMW \cite{MWRM}, whereas Section \ref{sec:methodCon} describes how these building blocks are utilized to obtain the \texttt{ExItNeRdoM} methodology. 

\subsubsection{Building blocks of the \texttt{ExItNeRdoM} methodology}
\label{sec:endetails}

\noindent The first task performed by each thread in the \texttt{ExItNeRdoM} (see Figure \ref{fig:diagram}) is uniform attribute binning with a width determined by Freedman-Diaconis rule \cite{Freedman1981OnTH}. It is denoted as \texttt{performBinning} in Algorithm \ref{alg:ExpInt}. This type of binning is used because it is necessary to account both for balanced and imbalanced data scenarios. Using probability  binning might mix a number of underrepresented classes into the same bin, which is not adequate for our use case. However, we do allow joining bins with high neuron activation values into a single bin if required to increase the bin size. Cases where bins contain small number of entities are also alleviated by the techniques of redesription mining that allow widening the bin interval if required. Assume we have an attribute $A$ with minimal value $v_{A,min}$ and a maximal value $v_{A,max}$ and that we obtained bins $[v_{A,min}, v_{A,t_1}], [v_{A,t_1},v_{A,t_2}]\dots, [v_{A,t_{k-1}},v_{A,max}]$. Each bin can be transformed in an analogue rule (query) of a form \texttt{$v_{A,t_{s-1}}\leq A\leq v_{A,t_{s}}$}. This is denoted as \texttt{createRules} in Algorithm \ref{alg:ExpInt}. Bin widening can, for example, be achieved by creating a query \texttt{$v_{A,t_{s-1}}\leq A\leq v_{A,t_{s}}\ \vee\ v_{A,t_{s}}\leq A\leq v_{A,t_{s+1}}$}. Such approach is denoted as \texttt{refine} in Algorithm \ref{alg:ExpInt}.  Bin joining can internally also be achieved through the use of negations, for example \texttt{$\neg (v_{A,min}\leq A\leq v_{A,t_{1}})$} describes entities from all bins except the first one. 

\begin{figure}[ht!]
\begin{center}
\includegraphics[width=0.8\textwidth]{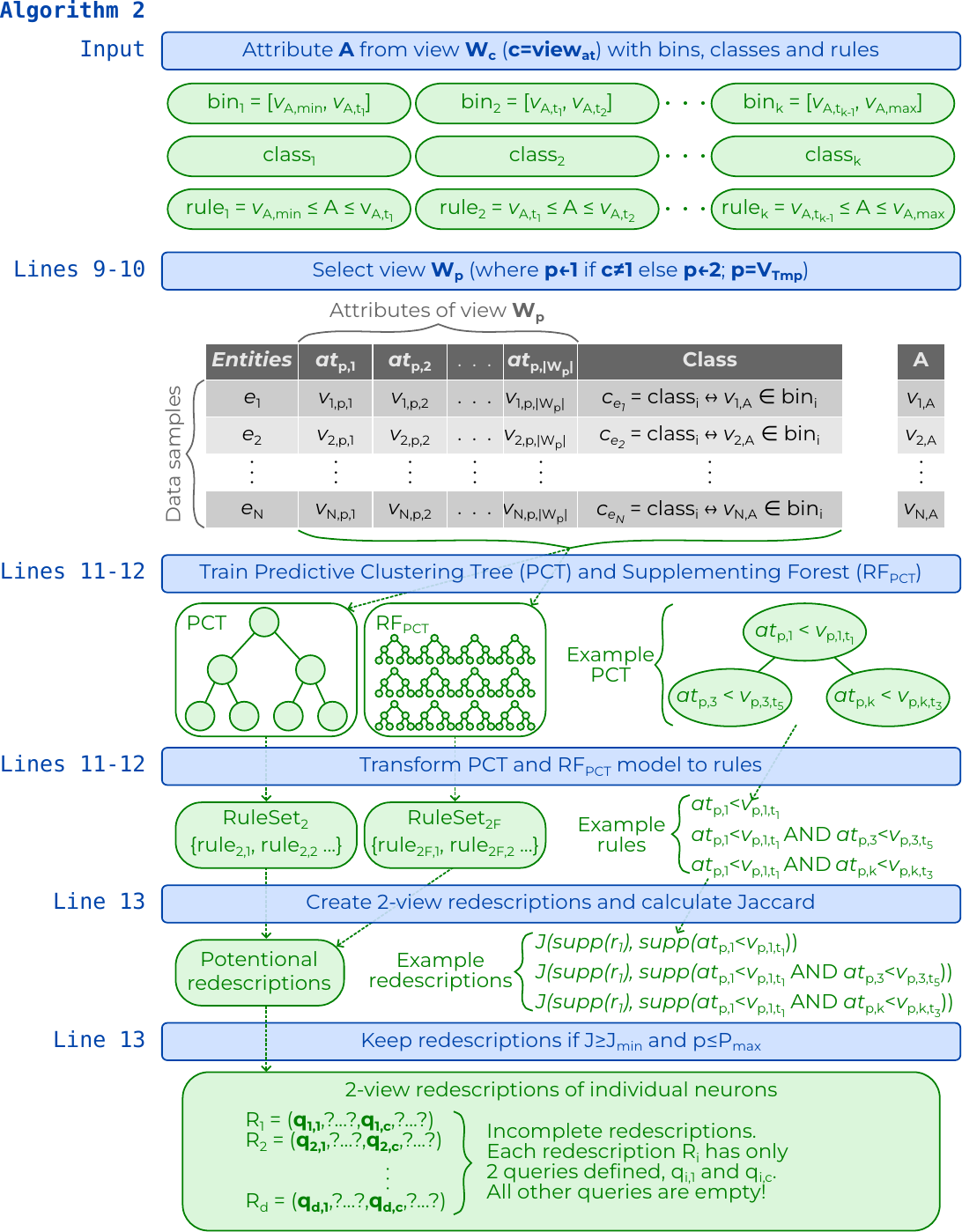}
\end{center}
\caption{Create $2$-view redescriptions. Bins of the input attribute $A$ from view $W_c$ are used to create classes of a multi-class classification problem ($class_i$) and rules ($rule_i$). Classes are used to
create a PCT and a supplementing random forest model on view $W_p$, $p\neq c$. The class value of entity $e_k$ equals $c_k$ if the selected attribute value of $e_k$ falls in the $k$-th bin. The obtained models are transformed to rules. Rules obtained from bins of $A$ and rules obtained from the models are used to create $2$-view redescriptions. Redescriptions with insufficient accuracy ($J<J_{min}$) and statistically insignificant redescriptions ($p>P_{max}$) are filtered out.}
 \label{fig:PCTInitRules}
\end{figure}

\begin{figure}[ht!]
\begin{center}
\includegraphics[width=0.8\textwidth]{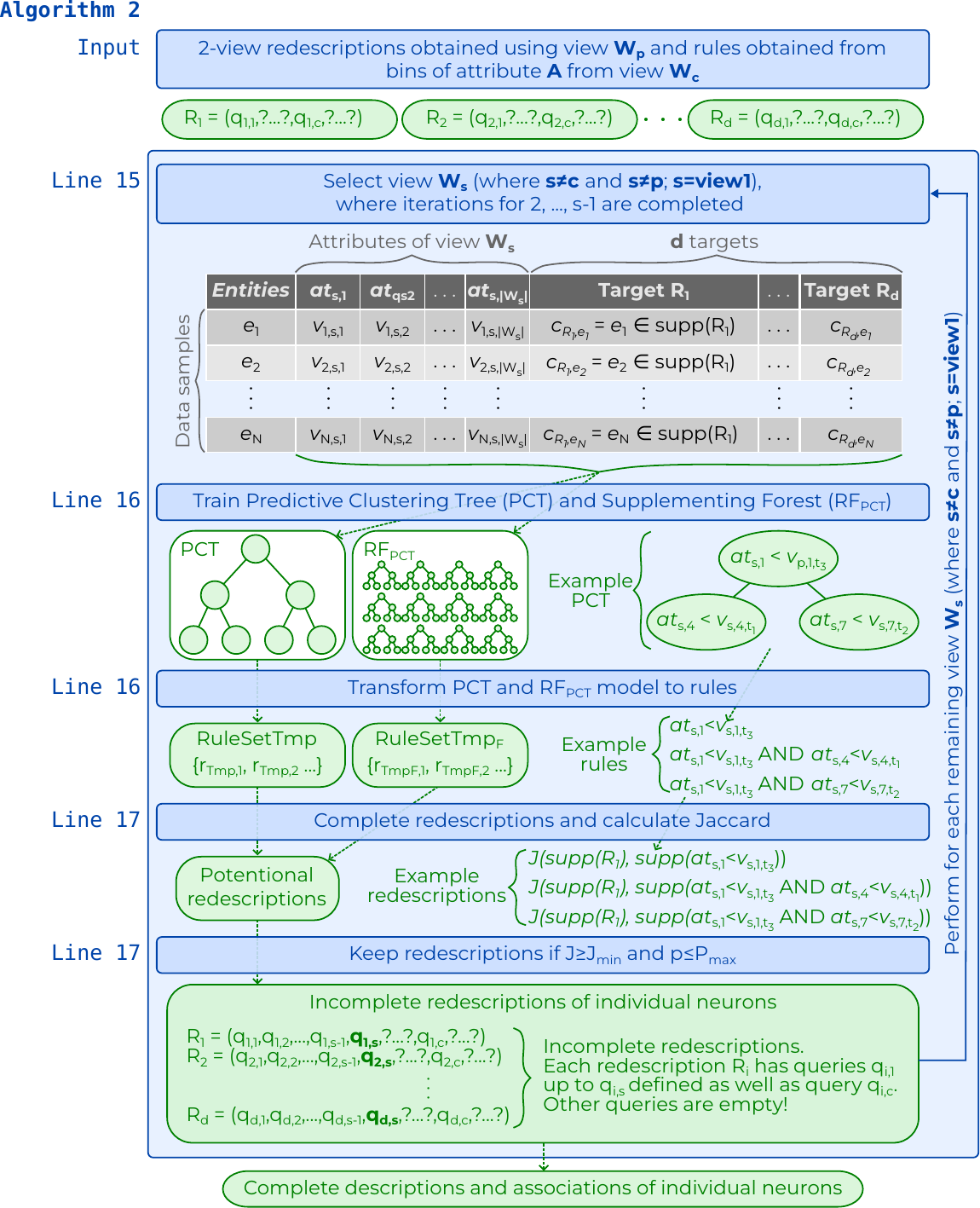}
\end{center}
\caption{Completing incomplete redescriptions. Input incomplete redescriptions are used as targets to create a PCT on view $W_s$, $s\neq c,\ p$. The $i$-th target for entity $e_k$ is $1$ if and only if $e_k\in supp(R_i)$. The obtained PCT is transformed to rules. Input rules are obtained from the PCT are used to complete the incomplete input redescriptions.}
 \label{fig:PCTRules}
\end{figure}

Rules obtained from the bins of some selected attribute  need to be related to the attributes contained in remaining input views. This is done in two phases: 
\begin{enumerate}
\item Predictive Clustering tree \cite{Kocev} (PCT) is trained on one of the remaining views, where each bin forms one class in a multi-class classification setting (only one target variable is constructed). The obtained PCT is transformed to rules, and the two sets of rules (a set obtained from attribute bins and a set obtained from the trained PCT) are used to create $2$-view redescriptions (see Figure \ref{fig:PCTInitRules}).  These two steps are denoted as \texttt{transformToRules} and \texttt{createReds} respectively in Algorithm \ref{alg:ExpInt}. The $2$-view redescriptions are incomplete if input data contains more than $2$ views and need to be completed in the following step. The procedure to create redescriptions using PCTs, where rules form target labels has been introduced in \cite{MihelcicRMRF}. The main modification in this work is using binning and rules obtained from attribute bins in redescription mining process, which allows obtaining redescriptions containing individual neurons and their activations.
\item PCTs are iteratively trained on all remaining views, so that redescriptions form target labels and the rules obtained from PCTs are used to complete redescriptions. The input dataset to train PCTs is constructed as in Figure \ref{fig:PCTRules}, but instead of placing $1$ in the $k$-th row for the $i$-th label if and only if $e_k\in supp(r_i)$, target labels are redescriptions, and the $k$-th row will have a value $1$ if and only if $e_k \in supp(R_i)$. Redescription completion is performed, so that $J(supp(R_i), supp(r_j))$ is maximized. This step is denoted as \texttt{completeReds} in Algorithm \ref{alg:ExpInt}. This procedure is internally used in the clustering redescription mining multi-view (CLUS-RMMW) algorithm \cite{MWRM}.
\end{enumerate}

  \begin{algorithm}[ht!]
\caption{The \texttt{CLUS-RMMWConst} algorithm.}\label{alg:CLUS-RMMWConst}
\begin{algorithmic}[1]
\Require{Selected neuron $n$, Input rule sets RuleSet$_1$, RuleSet$_2$, settings Set, domain data $\mathcal{D}$ with views $W_1$,\dots, $W_p$}
\Ensure{Redescription set containing redescriptions that have neuron $n$ occurance in a query constructed on a view containing this neuron.}
\Procedure{CLUS-RMMWConst}{}
\State [iter, $\mathcal{R}_{id(n),int}$] $\leftarrow [0, \emptyset]$
\State [$RuleSet_{W_1},\ RuleSet_{W_2}$]$\leftarrow$ [$RuleSet_1,\ RuleSet_2$]
\While{(iter$<$set.numIter)}
\State $RuleSet_{T_1}\leftarrow$ \texttt{transformToRules}(\texttt{PCT}($RuleSet_{W_1}$,$W_{view(A)}$)), \par \hspace{55mm} $A\in attrs(r), r\in RuleSet_{W_2}$
\State $RuleSet_{T_2}\leftarrow$ \texttt{transformToRules}(\texttt{PCT}($RuleSet_{W_2}$,$W_{view(n)}$))
\State $RuleSet_{FT_1}\leftarrow$ \texttt{transformToRules}($\texttt{RF}_{\texttt{PCT}}$($Set_{nt},\ RuleSet_{W_1}$, \par \hspace{45mm}$W_{view(A)}$)) $A\in attrs(r), r\in RuleSet_{W_2}$
\State $RuleSet_{FT_2}\leftarrow$ \texttt{transformToRules}($\texttt{RF}_{\texttt{PCT}}$($Set_{nt},\ RuleSet_{W_2}$,\par \hspace{9cm}$W_{view(n)}$))
\State $\mathcal{R}_{id(n),int}\leftarrow\ \mathcal{R}_{id(n),int}\ \cup\ $\texttt{createRedsConst}($n,\ RuleSet_{W_1},\ $ \par \noindent  $RuleSet_{T_1} \ \cup\ RuleSet_{FT_1},\ Set_{CRMSet},\ 1,\ max(view(A), view(n)),\ W_{v_{Tmp}}$)
\State $\mathcal{R}_{id(n),int}\leftarrow\ \mathcal{R}_{id(n),int}\ \cup\ $\texttt{createRedsConst}($n,\ RuleSet_{T_2}\ \cup $ \par \noindent  $RuleSet_{FT_2},\ RuleSet_{W_2},\ Set_{CRMSet},\ 1,\ max(view(A), view(n)),\ W_{v_{Tmp}}$)
\For{($view1\in Views\setminus \{view(A),\ view(n)\}$)}
\State $RuleSetTmp\leftarrow$ \texttt{transformToRules}(\texttt{PCT}($\mathcal{R}$, $W_{view1}$))
\State  $\mathcal{R}_{id(n),int}\leftarrow$ \texttt{completeReds}($\mathcal{R}_{id(n),int}$,$Set_{CRMSet}$,$view1$)
\EndFor
\State [$RuleSet_{W_1},\ RuleSet_{W_2}$]$\leftarrow$ [$RuleSet_{T_2},\ RuleSet_{T_1}$]
\EndWhile
\State \textbf{return} $\mathcal{R}_{id(n),int}$
\EndProcedure
\end{algorithmic}
\end{algorithm} 

To obtain the output redescription set describing interactions of a selected neuron with other neurons in the same or different layers of one or more DLMs, we adapt and extend the multi-view CLUS-RM approach \cite{MWRM} so that: 
\begin{itemize}
\item[a)] the algorithm takes additional inputs in the form of rules obtained from selected attribute binning and from the PCT used to obtain $2$-view redescriptions in the previous step (used to create descriptions of the selected individual neuron activations). This alleviates the need to perform initializations as done in  \cite{MWRM}. 
\item[b)] Since the goal is to obtain interactions of a selected neuron, not to discover a representative subset of all highly accurate redescriptions, it is not necessary to perform $2$-view redescription construction and completion process in a view-pairwise fashion as done in \cite{MWRM}. This significantly reduces the required time to obtain the resulting redescription set. 
\item[c)] We incorporate hard attribute constraints \cite{MihelcicADNI} in the modified multi-view redescription mining process to ensure the selected neuron is present in all discovered redescriptions, a feature unavailable in \cite{MWRM}. This is achieved by adding hard attribute constraints on rules used to create $2$-view redescriptions (only rules containing a selected neuron from a corresponding view are considered). Once $2$-view redescriptions contain a selected neuron in a corresponding query, these are naturally completed to multi-view redescriptions. This step further increases the overall speed of the performed multi-view redescription mining process. 
The constraint-based redescription mining \cite{MihelcicADNI} allows finding redescriptions from data containing only two views and does not allow creating separate redescription sets for different combinations of constraints or ensuring that redescription-analogous of individual neuron descriptions will be obtained. 
\end{itemize}

The procedure of constrained multi-view redescription mining, used to discover redescriptions containing interactions of a selected neuron, is fully described in Algorithm \ref{alg:CLUS-RMMWConst}. The algorithm iterates for a predefined number of iterations (line $4$) and creates targets and PCTs from each set of input rules in turn as described in Figure \ref{fig:PCTRules} (lines $5$ and $6$). Additionally, it can train a random forest (RF) containing a predefined number of $nt$ trees, as in \cite{MihelcicRMRF}, to increase the accuracy and diversity of obtained rules ($nt = 0$ denotes RF will not be used), see lines $7$ and $8$. The obtained rules are used to create $2$-view redescriptions containing attribute corresponding to the selected neuron $n$ (line $9$ and $10$). From the rules obtained on view $view(n)$, only these containing attribute corresponding to a selected neuron $n$ are used to construct redescriptions. $2$-view redescriptions are extended to multi-view (lines $11-13$) and the final set of redescriptions is returned as output (line $15$).  Note that RFs are not used to complete redescriptions, since it would require training a number of consecutive forest with many target variables (equal to the number of $2$-view redescriptions). In each iteration, the rules used to obtain PCTs are exchanged with rules obtained in the current iteration to enable space exploration (line $14$). This process is called alternation in the field of redescription mining (see \cite{RamakrishnanCart}).

\subsubsection{The \texttt{ExItNeRdoM} methodology}
\label{sec:methodCon}
The \texttt{ExItNeRdoM} methodology is described in Algorithm \ref{alg:ExpInt}. This algorithm takes as input a sequence of domain data views and views representing selected layers of one or more DLMs (see Fig. \ref{fig:overview}, part a, tables $W_i$). The first step in the computation process of the \texttt{ExItNeRdoM} methodology is to collect attributes from all views of interest and to divide them evenly across the available worker threads (lines $2$ and $3$). Lines $5-19$ are executed in parallel in each worker thread.  For each attribute assigned to the worker  thread (line $6$), we first perform binning (line $7$), using uniform binning with a width determined by Freedman-Diaconis rule (see Section \ref{sec:endetails} for details).



\begin{algorithm}[ht!]
\caption{The \texttt{ExItNeRdoM} methodology.}\label{alg:ExpInt}
\begin{algorithmic}[1]
\Require{Layers of DLMs $L_1^{\mathcal{M}_1},\dots,  L_{s_i}^{\mathcal{M}_1}, \dots, L_{s_k}^{\mathcal{M}_k}, \ s_i\geq 1,\ k\geq 1$ (or $\emptyset$ to emulate pedagogical rule extraction), domain data $\mathcal{D}$ views $W_1,\dots, W_p,\ p\geq 1$ (or $\emptyset$), target labels $\mathcal{Y} = (y_1,\dots ,y_c),\ c\geq 1$ (or $\emptyset$), settings \texttt{Set}.}
\Ensure{RS $\mathcal{F} = \{\mathcal{R}_1,\dots, \mathcal{R}_k\},\ k = 2\cdot (\sum_{i,j} |L_{i}^{\mathcal{M}_j}| + \sum_j |W_j| + |\mathcal{Y}|)$}
\Procedure{ExItNeRdoM}{}
\State $\mathcal{S}\leftarrow \cup_{i,k} attrs(L_{s_i}^{\mathcal{M}_k})\ \cup\  attrs (\mathcal{D})\ \cup \ $  $ attrs(\{y_1,\dots,y_c\})$
\State $[grSizes, grAttrs]\leftarrow$ \texttt{unifDiv}$(Set_{NumThreads},\ \mathcal{S})$
\State @do in parallel (thread$_i$)
\State $\mathcal{F}_i\leftarrow \emptyset$
\For{(at = 1; at$\leq grSizes[i]$; at++)}
\State $[view_{at},bins_{at},class_{at}]\leftarrow$ \texttt{performBinning}($grAttrs[i][at]$)
\State $InitRules\leftarrow$\texttt{createRules}($bins_{at}$)
\State $v_{Tmp}\leftarrow 1$
\If{($view_{at}==1$)} $v_{Tmp} = 2$
\EndIf
\State $RuleSet_2\leftarrow$ \texttt{transformToRules}(\texttt{PCT}($class_{at}$,$W_{v_{Tmp}}$))
\State $RuleSet_{2F}\leftarrow$ \texttt{transformToRules}($\texttt{RF}_{\texttt{PCT}}$($Set_{nt},\ class_{at}$,$W_{v_{Tmp}}$))
\State $\mathcal{R}_{id(grAttrs[i][at]),s}\leftarrow$\texttt{createReds}($InitRules,\ RuleSet_2\ \cup\ $\par \hspace{2cm} $RuleSet_{2F},\   Set_{CRMSet},\ 1,\ max(v_{Tmp}, view_{at}),\ W_{v_{Tmp}}$)
\State $\mathcal{R}_{id(grAttrs[i][at]),s}\leftarrow$ \texttt{refine}($bins, \mathcal{R}_{id(grAttrs[i][at]),s}$, $v_{Tmp}$, $W_{v_{Tmp}}$)
\For{($view1\in Views\setminus \{view_{at},\ v_{Tmp}\}$)}
\State $RuleSetTmp\leftarrow$ \texttt{transformToRules}(\texttt{PCT}($\mathcal{R}$, $W_{view1}$))
\State  $\mathcal{R}_{id(grAttrs[i][at]),s}\leftarrow$ \texttt{completeReds}($\mathcal{R}_{id(grAttrs[i][at]),s}$,\par \hspace{4cm} $set_{CRMSet}$,$RuleSetTmp$,$view1$)
\EndFor
\State $\mathcal{R}_{id(grAttrs[i][at]),int}\leftarrow$ CLUS-RMMWConst($grAttrs[i][at]$,  \par \hspace{32mm} $InitRules$, $RuleSet_2$,  $set_{CRMSet}$, $W_1, \dots, W_p$)
\State $\mathcal{F}_i\leftarrow \mathcal{F}_i\ \cup\ \{\mathcal{R}_{id(grAttrs[i][at]),s},\ $ $\mathcal{R}_{id(grAttrs[i][at]),int}\}$
\EndFor
\State $\mathcal{F}\leftarrow\ \cup_i\ \mathcal{F}_i$
\State \textbf{return} $\mathcal{F}$
\EndProcedure
\end{algorithmic}
\end{algorithm} 


\noindent  Each bin is transformed into a rule (line $8$). Depending on the view of the selected attribute, a Predictive Clustering tree (PCT) \cite{Kocev} is trained on view $0$ or $1$ using a multi-class classification dataset, where each bin forms one class (line $11$). 
The goal is to obtain a tree having similar groups of entities in its nodes as described by the aforementioned rules/classes. The obtained PCT is transformed into a set of rules (line $11$), which are in combination with rules obtained from bins used to create redescriptions (line $13$), see Section \ref{sec:endetails} and  \cite{MihelcicRMRF, MihelcicFramework}. Since we have a multi-view setting, the \texttt{createReds} function takes additional two parameters ($4$th and $5$th) denoting views on which input rule sets were created. It also uses $W_{viewTmp}$ view, needed for computation when all logical operators can be utilized in redescription query construction.  The obtained $2$-view redescriptions are extended so that a query containing rule obtained from some bin of a selected attribute is combined, using logical disjunction operator, with a rule corresponding to some other bin of this attribute, if it increases overall accuracy of this  redescription (line $14$). For example, redescription $R = (0.2 \leq n_{2,3}\leq 0.7,\ weather = Sunny)$ might be extended to $R' = (0.2 \leq n_{2,3}\leq 0.7\ \vee\ 0.8 \leq n_{2,3}\leq 0.9 ,\ weather = Sunny)$ if $J(R')\geq J(R)$. The obtained redescriptions are incomplete if input data contains more than $2$ views, since they contain only two queries (e.g. $R'' = (0.2 \leq n_{2,3}\leq 0.7,\ weather = Sunny,\ ?)$ ). If some layer is related to more than one view (layers of the same or different networks, or views of original data), the procedure iteratively creates PCTs on remaining views with redescriptions as targets and uses obtained rules to iteratively complete redescriptions (add corresponding queries) (lines $15-17$), see also explanations in Section \ref{sec:endetails}. The example $R''$ could potentially be completed to $R'' = (0.2 \leq n_{2,3}\leq 0.7,\ weather = Sunny,\ 10000\leq monthlyRevenue)$. 
Lines $13-17$ search for redescriptions that relate various bins of an attribute, including individual neuron activations, with attributes contained in other available views. However, often groups of neurons represent some useful object (e.g. a window, a cup in the image). To discover different neuron groups related to the selected attribute, the framework uses a modified multi-view CLUS-RM procedure \cite{MWRM} (see Section \ref{sec:endetails} for details), where initialization is replaced by rules obtained from lines $13-17$ and attribute constraints are added, following a constraint-based redescription mining paradigm \cite{MihelcicADNI}, to ensure the observed neuron occurs in every redescription. Thus, it searches for neuron and attribute combinations that accurately re-describe subsets of entities induced by attribute bins (potentially individual neuron activations or their refinement). In this way, the proposed framework discovers potentially interesting re-descriptions of individual neuron activations and of their interactions  with related groups of neurons (line $18$). 
The obtained redescription sets are added to a family of redescription sets by each thread (line $19$). Chunks of the resulting family are joined (line $20$) and returned as the output of the procedure.  

The \texttt{ExItNeRdoM} methodology incorporates supplementing Random Forest as in \cite{MihelcicRMRF} to increase rule number and diversity  (see line $12$) and inside the \texttt{CLUS-RMMWConst} (line $18$). The process consists of training the main PCT model and a supplementing forest, containing a predefined number of trees using identical target labels. 

\subsection{Utilizing outputs of the \texttt{ExItNeRdoM} methodology to perform rule extraction}

To perform rule extraction with the \texttt{ExItNeRdoM}, one must choose the views to be used to construct rules and redescriptions (original data, predictions made by the model, layers of the model). Additionally, a selection approach of the appropriate rules/redescriptions that describe predictions made by the chosen model must be constructed. With these additions, the \texttt{ExItNeRdoM} can emulate pedagogical approaches by taking the original input data as a first view and the predicted target labels from some DLM as the second view. Utilizing views describing layers of a DLM, the approach produces more advanced results (similar to \texttt{Eclair}, \texttt{DeepRED}). The  \texttt{ExItNeRdoM} outputs both rules, building blocks of redescriptions (as regular rule extraction algorithms), and redescriptions. Redescriptions can provide information about complex associations of neurons between layers, their relation to original attributes or predictions made by the chosen DLM. 

\begin{algorithm}[ht!]
\caption{Algorithm for explaining predictions of DLMs.}\label{alg:RSel}
\begin{algorithmic}[1]
\Require{Precision threshold $\delta$, DLM layers $L_1^{\mathcal{M}_t},\dots,  L_{s_i}^{\mathcal{M}_t}, \ s_i\geq 1$ (or $\emptyset$), domain data $\mathcal{D}$ views $W_1,\dots, W_p,\ p\geq 1$, predictions $\mathcal{Y}_p = (y_{1_p},\dots y_{c_p}),\ c\geq 1$, RS $\mathcal{F}$ from Algorithm \ref{alg:ExpInt} using $L_i^{\mathcal{M}_t}$, $\mathcal{D}$, $\mathcal{Y}_p$ and settings \texttt{Set}.}
\Ensure{Redescriptions $\mathcal{R}_o$, rules $\mathcal{R}_{r_o}$ with high fidelity to predictions of $\mathcal{M}_t$.}
\Procedure{ConstructSets}{}
\State \texttt{init}(\texttt{NcovEntReds}[c], \texttt{NcovEntRules}[c], $|E|_c$), $\forall c\in $\texttt{classValue}
\State \texttt{init}(\texttt{chosenReds}[c], \texttt{chosenRules}[c], $\emptyset$), $\forall c\in $\texttt{classValue}
\State \texttt{init}($\mathcal{R}_{sel,c},\ \mathcal{R}_{rsel,c},\ \emptyset$), $\forall c\in $ \texttt{classValue}
\For{(index = 0; index < $|classValue|$; i++)}
\State \{\texttt{max, maxR}\} $\leftarrow \{-1, \emptyset$\}
\For{($R\in \mathcal{F}$)}
\If{(precision($R$) $<\delta$)} \texttt{continue}
\EndIf
\State \{\texttt{numCov}, \texttt{score}\}$\leftarrow 0$
  \If{(\texttt{NcovEntReds[index]$>0$})}
  \State \texttt{numCov} $\leftarrow$ \texttt{countCovered($R$)}
  \State \texttt{score} $\leftarrow $  ((\texttt{numCov}$/$\texttt{NcovEntReds[index]}) $+$ accuracy($R$) $+$ \par \hspace{3cm}$2\cdot$ precision($R$))/$4$
  \If{(score>max)} $\{max,\ maxR\}$ $\leftarrow$  $\{score,\ R\}$
  \EndIf
  \State \texttt{chosenReds[index]}$\leftarrow$ \texttt{chosenReds[index]}$\cup$ maxR
  \State \texttt{NcovEntReds}[c]$\leftarrow$ \texttt{NcovEntReds}[c] - numCov
  \Else
  \For{($R_c \in$ \texttt{chosenReds[index]})}
  \State \texttt{equalSupp}$\leftarrow$ \texttt{isEqual($R_c, R$)}
  \If{(\texttt{equalSupp} == 1 $\wedge$ $R.JS>R_c.JS$ $\wedge$ \texttt{precision($R$)} \par \hspace{6cm}$\geq$ \texttt{precision($R_c$)})}
  \State \texttt{chosenReds[index]}$\leftarrow$ \texttt{chosenReds[index]}$\setminus$ $R_c\ \cup R$
  \EndIf
  \EndFor
  \EndIf
\EndFor
\EndFor
\State repeat lines $5-20$ for rules
\State \textbf{return} \{\texttt{chosenReds}, \texttt{chosenRules}\}
\EndProcedure
\end{algorithmic}
\end{algorithm} 
\noindent The \texttt{ExItNeRdoM} can be applied as is to multi-label or multi-target scenarios, where it is possible to include various interactions between different labels. 

The presented optimization approach selects redescriptions based on redescription Jaccard index (accuracy), precision of mimicking input model predictions and entity coverage, whereas rules are selected based on entity coverage and precision of mimicking model predictions, explained in Algorithm \ref{alg:RSel}. Rule score (line $12$ analogue) is computed as: $(\texttt{numCov}/\texttt{NcovEntReds}[index] + 2\cdot \texttt{precision}(R))/3$.
 The approach can be additionally improved by the use of submodular optimization \cite{YangHYDYYS21}. Algorithm \ref{alg:RSel} is applied to every fold of the data. After redescription and rule sets are obtained, the fidelity score \cite{zarlenga2021efficient} is computed on each fold for the model used to obtain predictions. 
 The optimization process first describes all entities with redescriptions of high precision and accuracy, trying to use as small number as possible (lines $5 - 15$). After all entities are described, it iteratively replaces existing candidates with better available candidates (lines $16 - 20$). The same procedure is repeated for rules (line $21$) and the best obtained candidates are returned to the user (line $22$). 

\section{Complexity analyses and scalability study of the \texttt{ExitNeRdoM}}
\label{sec:complAndScal}
We first present the theoretical analyses of the time complexity of the approach. Next, we perform a scalability study with respect to the most influential parameters, the number of entities in the dataset, the number of neurons present in a selected layer of the DLM and the number of DLM layers being explained. 

\subsection{Complexity analyses}
We will first analyse time complexity of computation performed by each thread. The result will be used to compute time complexity of the \texttt{ExitNeRdoM}. 

Attribute binning of a selected node $n$, performed as a starting step of computation, has the worst time complexity of $\mathcal{O}(|E|\log_2|E|)$ if all attribute values are distinct. Bin conversion to rules has a worst-time complexity of $\mathcal{O}(k\cdot |E|)$, where $k$ equals the number of obtained bins. Given that $k\sim |E|^{1/3}$ \cite{Freedman1981OnTH}, this step has a worst-time complexity of  $\mathcal{O}(|E|^{4/3})$. Creating class labels from bins has equal worst-time complexity of $\mathcal{O}(|E|^{4/3})$. The time complexity of training a PCT, used to obtain matching rules to rules obtained from bins, is $\mathcal{O}(|V_i|\cdot |E|\cdot \log_2^2|E|) + \mathcal{O}(|V_i|\cdot |E|\cdot \log_2|E|) + \mathcal{O}(|E|\cdot \log_2 |E|)$, where $|V_i|$ denotes the number of attributes in the $i$-th view and $i = 0$ if $view(n) \neq 0$, otherwise $i = 1$ (see \cite{Kocev}). Transforming this PCT to rules has a complexity $\mathcal{O}(z)$, where $z$ equals the number of internal nodes in this tree. Equal complexity applies to training a Random Forest of PCTs if performed (the training procedure is carried number of trees times, which is a constant). Creating $2$-view redescriptions has an average time complexity of $\mathcal{O}(z^2\cdot |E|)$ and the worst time complexity, in case of inadequate hashing, of $\mathcal{O}(z^2\cdot |E|^2)$ (see \cite{MihelcicRMRF}. In practice, redescription support is smaller than $|E|$ and mostly limited to be $\leq c\cdot |E|$, for some $0<c<1$. If a technique of redescription refinement is used (see \cite{MihelcicFramework}) the complexity can increase maximally to $\mathcal{O}(z^4\cdot |E|)$ at average, although this scenario is unlikely due to internal constraints placed on redescriptions. Training PCTs to obtain rules to complete $2$-view redescriptions to multi-view redescriptions has complexity of $\mathcal{O}(\sum_{j\notin \{view(n),\ i\}} (|V_j|\cdot |E|\cdot \log_2^2|E|) + (|\mathcal{R}|\cdot |V_j|\cdot |E|\cdot \log_2|E|) + (|E|\cdot \log_2 |E|))$, this is smaller or equal to  $\mathcal{O}((|\mathcal{W}|-2)\cdot ((|V_{ks}|\cdot |E|\cdot \log_2^2|E|) + (|\mathcal{R}|\cdot |V_{ks}|\cdot |E|\cdot \log_2|E|) + (|E|\cdot \log_2 |E|)))$, where $ks = \text{argmax}_j |V_j|,\ j\notin \{view(n),\ i\}$. It should be noted that the maximal size of a redescription set $|\mathcal{R}|$ is bounded, as well as the maximal number of PCT targets (see \cite{MWRM} for more details). The average time complexity of completing $2$-view redescriptions is $\mathcal{O}(z\cdot |\mathcal{R}|\cdot |E|)$, since maximally $z$ rules obtained from a PCT in each step are used to try to complete each of $|\mathcal{R}|$ redescriptions. The worst time complexity, in case of inadequate hashing, is $\mathcal{O}(z\cdot |\mathcal{R}|\cdot |E|^2)$. As with rules, redescription support is smaller than $c\cdot |E|$ for $0<c<1$.  The overall average time complexity of obtaining redescriptions describing selected neuron and its activations is $\mathcal{O}(z^2\cdot |E| + |V_i|\cdot |E|\cdot \log_2^2|E| + |V_i|\cdot |E|\cdot \log_2|E| + |E|\cdot \log_2 |E| + (|\mathcal{W}|-2)\cdot ((|V_{ks}|\cdot |E| \cdot \log_2^2|E|) + (\mathcal{R}\cdot |V_{ks}|\cdot |E|\cdot \log_2|E|) + (|E|\cdot \log_2 |E|) + z\cdot |\mathcal{R}|\cdot |E|))$. The number of internal nodes of a PCT $z$ is bounded since the maximal PCT depth is bounded (usually to $6$), the redescription set size is also bounded (usually to $10 000$ candidates) and the number of views $\mathcal{W}$ is usually counted in tens, hundreds at most (even if someone would like to relate all layers of a ChatGPT4 - a task requiring ultimate computing power, it would take $\approx 120$ views). This gives the average complexity of $\mathcal{O}(|V_i|\cdot |E|\cdot \log_2^2|E| + (|\mathcal{W}|-2)\cdot (|V_k|\cdot |E| \cdot \log_2^2|E|) + |E|^{4/3})$.

The final step is to describe interactions of a selected neuron with other neurons from the same or other views. The $2$-view redescriptions are obtained by training a sequence of PCT pairs for a predefined number (\texttt{numIter}) of iterations. The complexity of this step is $\mathcal{O}(2\cdot numIter \cdot (|V_s|\cdot |E|\cdot \log_2^2|E|) + \mathcal{O}( |V_s|\cdot |E|\cdot \log_2|E|) + \mathcal{O}(|E|\cdot \log_2 |E|))$, where $s = \text{argmax}_{\substack{i\in \{view(n),0|\ view(n) \neq 0\}, \\ i\in \{0,1| view(n) = 0\}} }|V_i|$. Transforming PCTs to rules has complexity $\mathcal{O}(z)$ and $2$-view redescription creation has a complexity of $\mathcal{O}(z^2\cdot |E|)$. Taking into account constraints as above, this gives an average complexity of $\mathcal{O}(|V_s|\cdot |E|\cdot log_2^2 |E|)$. The complexity of completing $2$-view redescriptions is as above $\mathcal{O}((|\mathcal{W}|-2)\cdot ((|V_{ks}|\cdot |E|\cdot \log_2^2|E|) + (|\mathcal{R}|\cdot |V_{ks}|\cdot |E|\cdot \log_2|E|) + (|E|\cdot \log_2 |E|)))$, where $ks = \text{argmax}_j |V_j|,\ j\notin \{view(n),\ r\}, r = 0|view(n)\neq 0, \ r = 1|view(n) = 0$. This gives the average complexity of $\mathcal{O}(|V_s|\cdot |E|\cdot \log_2^2|E| + (|\mathcal{W}|-2)\cdot (|V_{ks}|\cdot |E| \cdot \log_2^2|E|))$.

The previously described procedure is repeated for every neuron of interest and is uniformly divided to $nt$ threads. Thus, the overall average complexity of the approach is: 
$\mathcal{O}(\frac{\sum_{c\in G}|V_c|}{nt}\cdot (|V_s|\cdot |E|\cdot \log_2^2|E| + (|\mathcal{W}|-2)\cdot (|V_{ks}|\cdot |E| \cdot \log_2^2|E|) + |E|^{4/3}))$, where $G$ denotes the subset of selected views of interest.
\subsection{Scalability study}
\label{sec:scalStudy}
In this section we study execution times of the \texttt{ExitNeRdoM} methodology dependent on: a) the number of entities in the dataset, b) the number of attributes in the dataset and c) the number of views in the dataset. We use the same algorithmic parameters as used for experiments presented in Section \ref{sec:expreldl}. All used parameters are specified in the supplementary document accompanying this manuscript. 

\subsubsection{Scalability with respect to the number of entities}
\label{sec:analEnt}
We performed the entity-based scalability study by relating penultimate layers of two multilayer perceptron networks initialized with different random initializations and trained on the MNIST dataset. Full details of the architecture are available in the supplementary document.

\begin{figure}[ht!]
\centering
\includegraphics[width=0.8\textwidth]{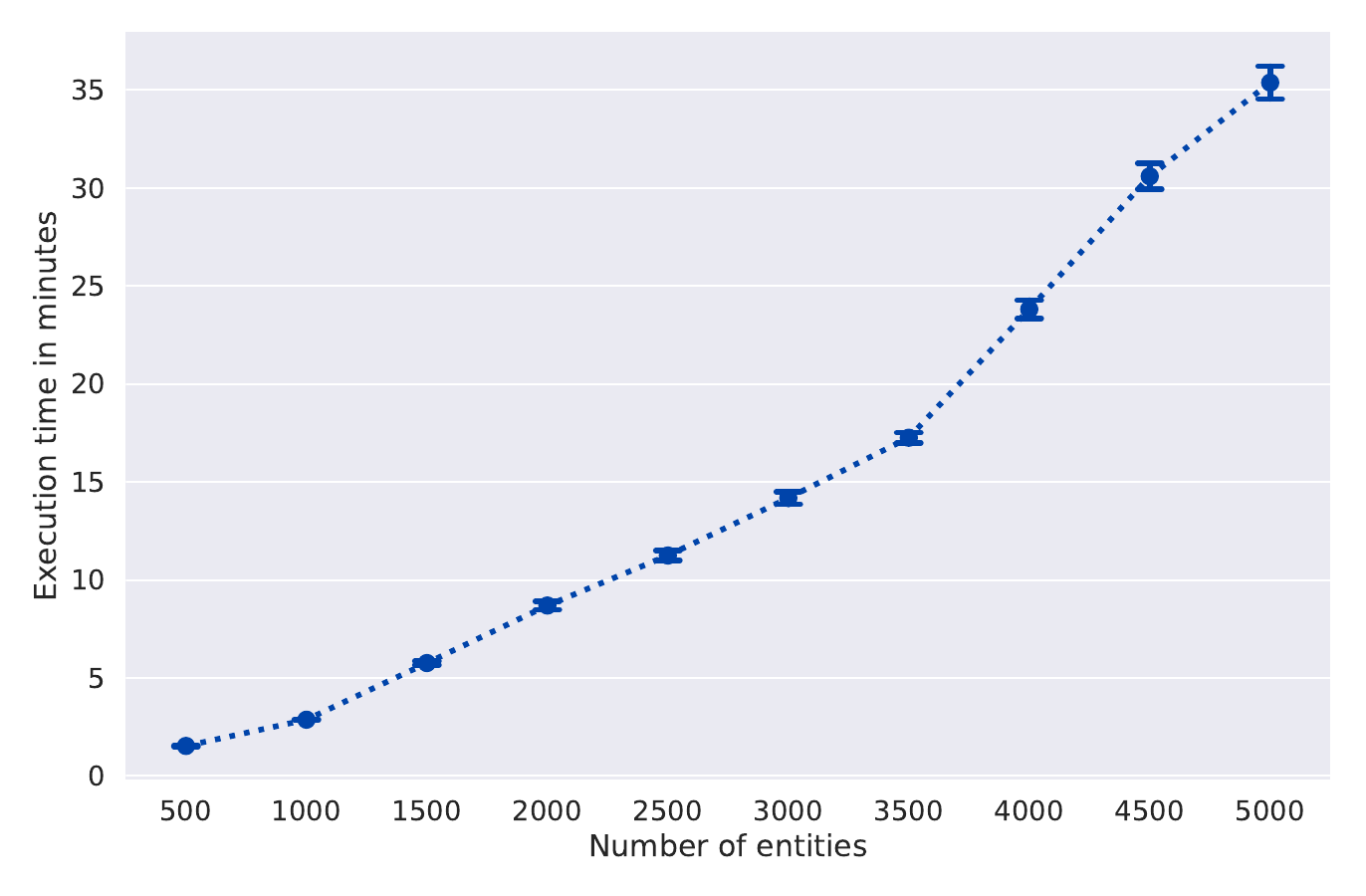}
\caption{Execution times in minutes of the \texttt{ExitNeRdoM} methodology required to relate penultimate layers of two multilayer perceptron networks. Each network contains $30$ neurons in the penultimate layer, thus $60$ neurons are described in total. Experiments are repeated $10$ times to measure potential variability due to use of a supplementing random forest.} 
 \label{fig:etEnt}
\end{figure}

\noindent The penultimate layer of each network contains $30$ attributes, thus $60$ attributes are described in total. The number of entities is varied between $500$ and $5000$ with an increment step of $500$. The execution time in minutes is shown in Figure \ref{fig:etEnt}. We repeated each experiment $10$ times and report the mean execution time (the point in the plot), and the bars represent deviations in execution times. Variability occurs due to the use of supplementing model of random forest, which selects random subsets of attributes during construction. 

As it can be seen from Figure \ref{fig:etEnt}, the use of supplementing random forest does not cause large variability in execution times. On $1000$ - $3500$ entity experiments, the framework shows linear increase in execution times (adjusted $R^2$ of $0.999336$). For larger numbers of entities, the constants increase due to the larger number of discoverable patterns, with the execution time being closer to the theoretically predicted boundary. 

\subsubsection{Scalability with respect to the number of neurons}
\label{sec:analAtt}

To perform the scalability study based on the number of neurons available in the dataset, we created a pair of multilayer perceptron neural networks with $250$ neurons in the penultimate layer. Each network was initialized with different random initialization. We used $500$ entities to perform all experiments. First, we selected the first $50$ attributes (neurons) of each network and related them using the \texttt{ExitNeRdoM} methodology. In consecutive steps, we used all the attributes from the previous step and added next $50$ attributes of each network.

\begin{figure}[ht!]
\centering
\includegraphics[width=0.8\textwidth]{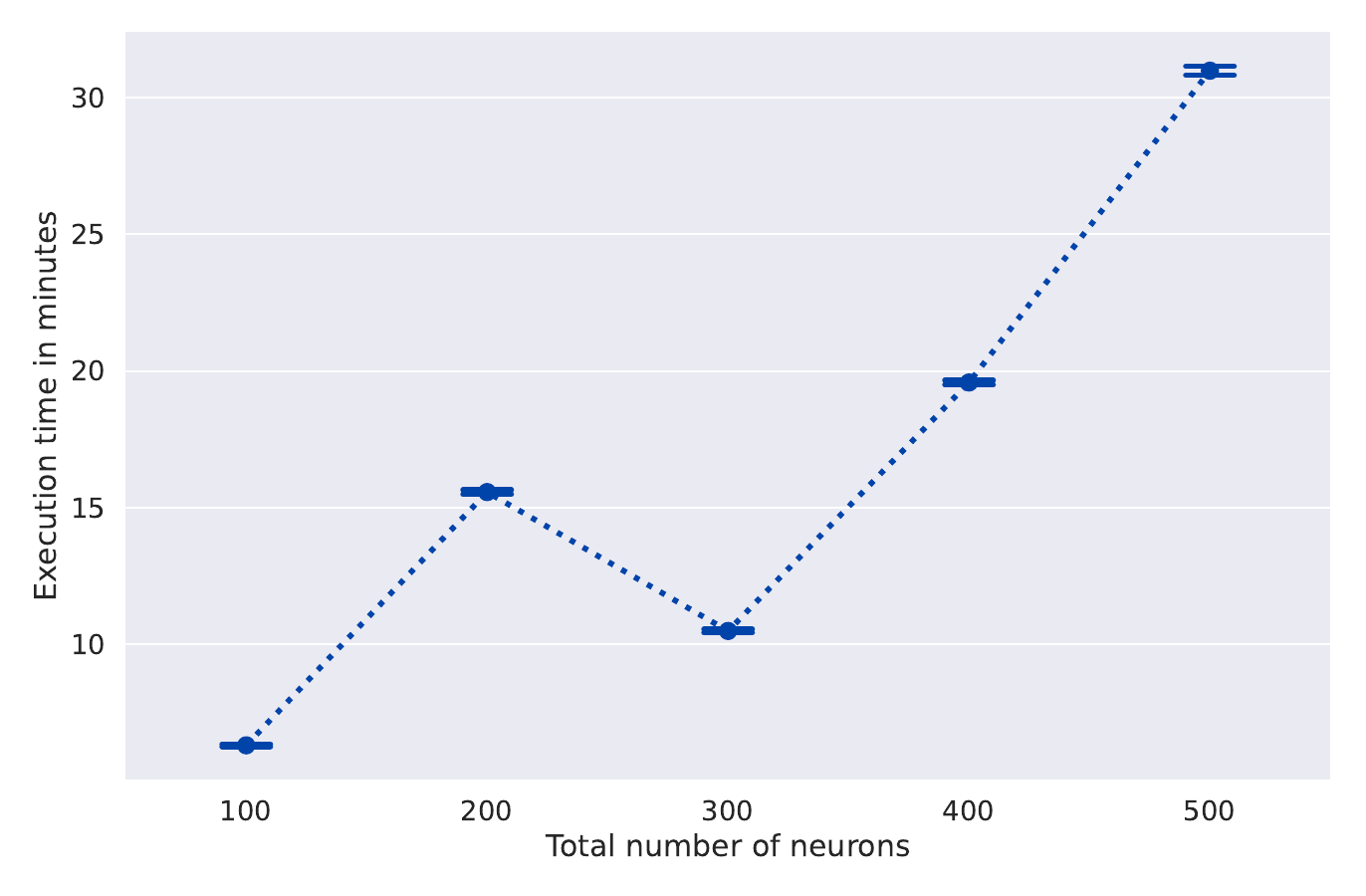}
\caption{Execution times in minutes of the \texttt{ExitNeRdoM} methodology required to relate penultimate layers of two multilayer perceptron networks on data containing $500$ entities. The first experiment was performed using first $50$ out of $250$ neurons of each network ($100$ neurons are described in total). At each consecutive experiment, the next $50$ attributes from each network are added to the set of attributes used in previous analyses. The $x$-axes shows the total number of described neurons from both networks.  Experiments are repeated $10$ times to measure potential variability due to use of a supplementing random forest.} 
 \label{fig:etAtt}
\end{figure}

 \noindent The total number of neurons being described is $100$, $200$, $300$, $400$ and $500$. The aforementioned experimental setup is  used to alleviate the impact of network architecture on the execution times, since the same neurons, although with partial connectivity, are used in each experiment. 

Experiments from Figure \ref{fig:etAtt} demonstrate that the variability of execution times caused by the use of random forest supplementing model is again very small. The increase in execution time is at most quadratic, as indicated by the theoretical analyses, with the observation that the last three points have a strong linear tendency (adjusted $R^2 = 0.9915221$). In the experiments, all neurons contained in the penultimate layer of both networks are described (thus $G = W$). The anomaly with $100$ neurons per network potentially occurred due to the fact that a part of neurons of the original network are missing, causing underlying trees to use a number of lower quality candidates to describe them. This would cause trees to have more branches and would ultimately lead to more rules, which increases the constants.

\subsubsection{Scalability with respect to the number of network layers}
\label{sec:analViews}

\begin{figure}[ht!]
\centering
\includegraphics[width=0.8\textwidth]{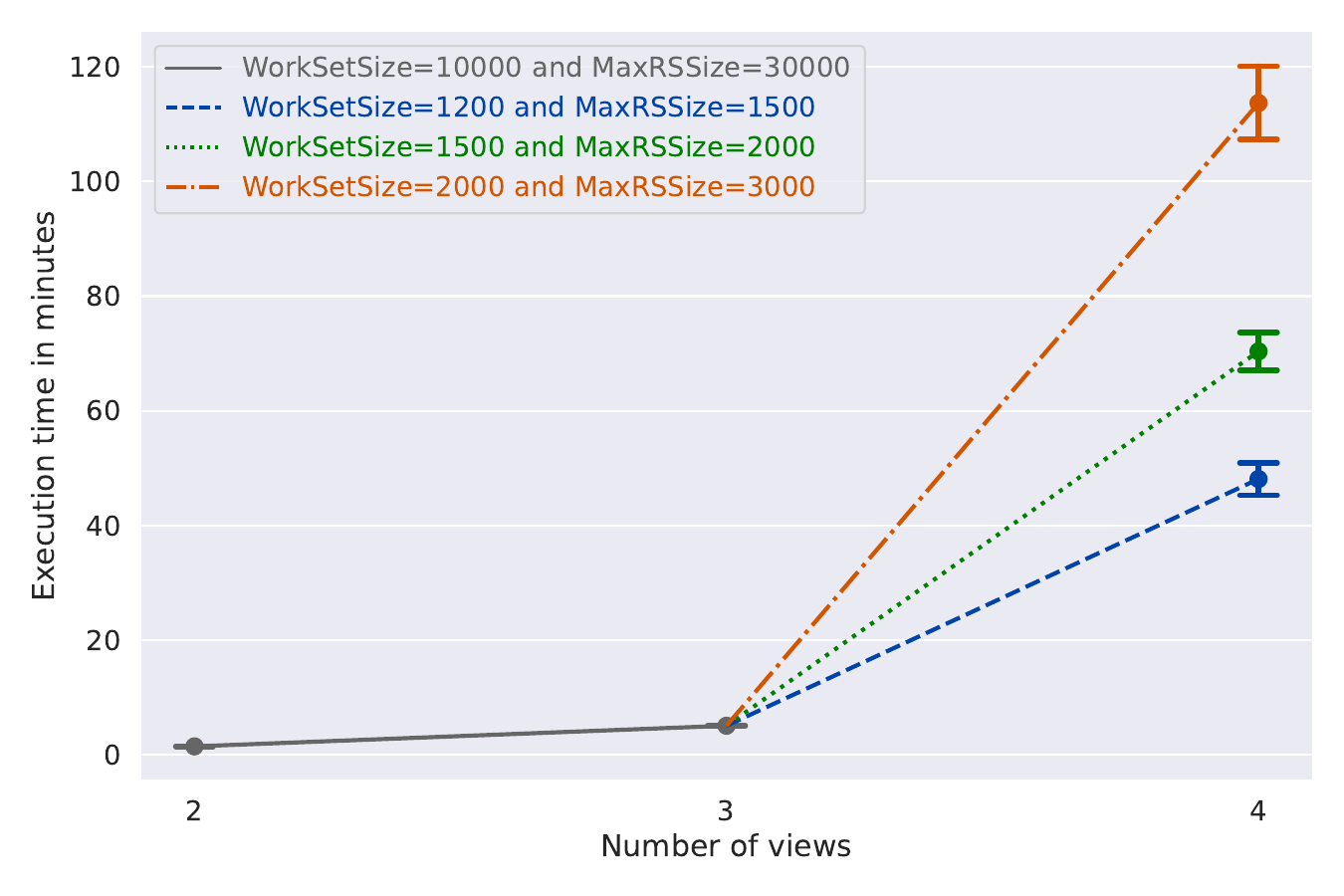}
\caption{Execution times in minutes of the \texttt{ExitNeRdoM} methodology required to relate penultimate layers of two, three and four multilayer perceptron networks on data containing $500$ entities and $30$ attributes per network. The $2$-view and $3$-view experiments were performed with \texttt{WorkSetSize} $=10000$ and  \texttt{MaxRSSize} $= 30000$, whereas the $4$-view experiments are performed with  (\texttt{WorkSetSize}, \texttt{MaxRSSize} ) settings: $(1200, 1500)$, $(1500, 2000)$ and $(2000, 3000)$. As it can be seen, increasing the size of \texttt{WorkSetSize} and  \texttt{MaxRSSize} parameters significantly increases the execution time of the approach. This is the expected behaviour since a larger number of candidates is held in memory. As a consequence, the time required to construct the final optimized set of redescriptions increases - since final redescriptions are extracted from a larger set of candidates, the overall number of produced rules increases - since attempts are made to complete or improve all candidate redescriptions, and the time required to construct new redescriptions, perform iterative candidate improvements and the query size reduction increases. However, using larger candidate sets can lead to overall better redescription sets. 
} 
 \label{fig:etView}
\end{figure}

Here, we study the impact of the number of related DNN layers to the execution times of the \texttt{ExitNeRdoM} methodology. The increase in the number of layers can cause an exponential increase in the number of redescriptions of discoverable patterns. Every $2$-view redescription can potentially be extended in multiple ways to form a $|\mathcal{W}|$-view redescriptions. Theoretically, from each $2$-view redescription, one may obtain ${|V| \choose k}^{|\mathcal{W}-2|}$ $|\mathcal{W}|$-view redescriptions, assuming each view contains $|V|$ attributes, and maximally $k$ attributes can be used to form a redescription query. The \texttt{ExitNeRdoM} adopts the techniques from multi-view redescription mining \cite{MWRM} to combat this increase. 
Internally, a number of redescription candidates are stored in memory, from which the final optimized set of redescriptions is created. 
The \texttt{WorkSetSize} parameter defines the number of top candidates that are stored during the entire execution of the approach. The \texttt{MaxRSSize} parameter defines the maximal number of redescriptions that can be kept in memory. Thus, the set containing \texttt{WorkSetSize} can be extended in iterations to obtain better candidates or to increase the quality of existing candidates. After each iteration, the excess candidates are removed from memory. As expected, both parameters have an impact on the execution time of the approach.  

We perform the $2$-view and $3$-view experiments using \texttt{WorkSetSize} $=10000$ and 
\texttt{MaxRSSize} $= 30000$. On the $4$-view experiments we study three different (\texttt{WorkSetSize}, \texttt{MaxRSSize} ) settings: $(1200, 1500)$, $(1500, 2000)$ and $(2000, 3000)$. Each experiment is repeated $10$ times to study potential variability in execution times caused by the use of supplementary random forest model. The experimental results can be seen in Figure \ref{fig:etView}.

\section{Experiments}
\label{sec:exp}
The experiments are divided in three groups. The aim of the first group (Section \ref{sec:rt}) is to determine whether the proposed methodology describes more neurons when relating pairs of DNNs with larger degree of similarity. We analyse the number of successfully described neurons by the \texttt{ExitNeRdoM} when relating pairs of ANNs with randomly assigned weights and compare this to corresponding pairs of trained ANNs. Next, we compute the correlation between obtained numbers of successfully described neurons to the state of the art measures for computing similarity between neural network representations, the CKA \cite{CKA} and the SVCCA \cite{SVCCA}. In the second group of experiments (Section \ref{sec:expdl}), we evaluate the performance of the proposed approach as a surrogate model that explains predictions made by DLMs and compare it with the state of the art rule extraction models. Finally in the third group (Section \ref{sec:expreldl}), we present the application of the \texttt{ExitNeRdoM} as XAI tool with a sequence of experiments that relate different number of DLMs of various type. The approach is compared to the original redescription mining approaches: the ReReMi \cite{GalbrunBW} which is utilized in the methodology by M\'{e}ger \cite{RmForDL} and the multi-view GCLUS-RM \cite{MWRM}. Information about utilized DLMs, and utilized parameters of the \texttt{ExitNeRdoM} methodology, are explained in the Supplementary document. 

\subsection{Randomization tests}
\label{sec:rt}

We conducted randomization tests on three different datasets, the tabular ADNI \cite{adni}, the Breast Cancer Wisconsin \cite{bcdata} and the image CIFAR-10 datasets \cite{krizhevsky2009learning}. On the ADNI and Breast Cancer Wisconsin datasets, we use Multilayer Perceptrons. On the CIFAR-10 dataset, we use a variant of the Convolutional Neural Network model called the ResNet-18 model \cite{ResNet}. 


We selected $10$ pairs of ANNs with randomly assigned weights and $10$ pairs of corresponding trained ANNs. The layer before the last (prediction) layer (the penultimate layer) was used to relate neuron activations in these layers, within the selected pair of ANNs with randomly assigned weights and within the corresponding pair of trained ANNs. $20$ pairs of ANNs were studied for each dataset. The test dataset of ANNs is used to relate neurons of resulting networks. We counted the number of neurons described by the approach for each pair of networks with randomly assigned weights and each pair of trained networks and computed the statistical difference of the median of the resulting numbers between trained and not-trained networks using Mann-Whitney U test. We also computed Spearman's correlation coefficient between the numbers of described neurons and the similarity scores between chosen layers of pairs of networks obtained using the linear kernel CKA \cite{CKA} and the SVCCA \cite{SVCCA}. 

To assess the difference in median value of the number of highly accurately described nodes of aforementioned ANNs, we use minimal Jaccard index of $0.8$, maximal $p$-value of $0.01$, minimal support of $10$ and maximal support of $0.8\cdot |E_{test}|$ (the last three parameter values are standardly used, see \cite{MihelcicRMRF}). Datasets on which the \texttt{ExitNeRdoM} was applied contained $275$ entities and $24$ neurons per network ($48$ in total) on the ADNI dataset, $171$ entities and $64$ neurons per network ($128$ in total) on the WDBC dataset and $1000$ entities and $512$ neurons per network ($1024$ in total) on the CIFAR-10 dataset.
We utilize only conjunction operator in redescription query construction to obtain focused, easily understandable knowledge that uncovers strong associations between neurons from a chosen pair of DLMs. 

The performed experiments demonstrate that the approach manages to re-describe neurons and their interactions of trained pair of DLMs with significantly higher median of re-described individual nodes (corresponding $p$-value denoted with $p_1$), interactions (corresponding $p$-value denoted with $p_2$) and total number of nodes + interactions (corresponding $p$-value denoted with $p_3$) compared to matching DLMs with randomly assigned activations. The corresponding results can be seen in Table \ref{tab:RRand}. This confirms the intuition that ANNs trained on the same set of target labels, even when different initializations are used, encode the relevant information in their neurons (potentially in physically different neurons using groups of neurons of different sizes). The proposed approach is able to detect and describe these similarities, which leads to creation of significantly larger number of highly accurate redescriptions compared to the number present in ANNs with random activations. Similarities within pairs of ANNs with random activations are rare and occur by chance. 
The corresponding correlation computation between the number of individually described neurons (in) and the CKA and SVCCA scores shows these correlations to be significant in majority of cases with only $10$ measurements. $SVCCA_{10}$ and $SVCCA_{20}$ denote the SVCCA calculated on $10$ and $20$ largest eigenvalues, respectively. CKA was calculated with linear kernel. The obtained correlations can be seen in Table \ref{tab:RRand}.  This experiment shows that the number of functional similarities of individual neurons, discovered by the proposed approach, significantly correlates with perceived similarity between pairs of studied ANNs, which is the desired behaviour of the methodology. Correlation with CKA on the CIFAR-10 dataset is noticeable, although not statistically significant. Increasing the number of measurements might resolve this anomaly.

\begin{table}
\caption{Significance of difference in median value of the number of highly accurately described nodes of the trained ANNs compared to ANNs with random activations (top $3$ rows). Significance for individual nodes is denoted $p_1$, interactions $p_2$ and total $p_3$. Bottom $3$ rows provide information on significance of Spearman correlation between the number of individually described neurons and ANN similarity measures CKA and SVCCA.}
\label{tab:RRand}
\begin{center}
\begin{tabular}{ |c|c|c|c| } 
\hline
Measure & ADNI & BreastCancer  & CIFAR-10 \\
\hline
$p_1$ & $8.68\cdot 10^{-5}$ & $8.93\cdot 10^{-5}$  &  $8.53\cdot 10^{-5}$  \\ 
$p_2$& $8.44\cdot 10^{-5}$ & $8.93\cdot 10^{-5}$ &  $8.93\cdot 10^{-5}$ \\ 
$p_3$& $8.58\cdot 10^{-5}$ & $8.88\cdot 10^{-5}$ &  $9.08\cdot 10^{-5}$ \\
$SP_{in,CKA}$& $1.5\cdot 10^{-5}$ & $9.82\cdot 10^{-6}$ & $0.16$   \\
$SP_{in, SVCCA_{10}}$& $2.7\cdot 10^{-4}$ & $3.6\cdot 10^{-4}$ &  $5.8\cdot 10^{-6}$ \\
$SP_{in, SVCCA_{20}}$& $3.9\cdot 10^{-4}$ & $4.1\cdot 10^{-4}$ &  $1.6\cdot 10^{-6}$ \\ 
\hline
\end{tabular}
\end{center}
\end{table}



\subsection{Explaining predictions of DLMs using \texttt{ExItNeRdoM}}
\label{sec:expdl}
We evaluate the feasibility of performing rule extraction task with the \par \noindent \texttt{ExItNeRdoM} and compare our approach with state of the art rule extraction approaches: \texttt{Eclair} \cite{zarlenga2021efficient}, \texttt{Rem-T} \cite{shams2021rem}, \texttt{Rem-D} \cite{shams2021rem}, \texttt{DeepRED} \cite{zilke2016deepred} and \texttt{PedC5.0} pedagogical approach \cite{kola}. The task was performed on $2$ Multilayer Perceptron ANNs, one trained on the ADNI dataset and the other on the Breast Cancer Wisconsin dataset. Both models contain an input layer, $2$ hidden layers and an output layer ($4$ layers in total). 

\subsubsection{Parameter settings and results}

Algorithm \ref{alg:ExpInt} was used to obtain rules and redescriptions, minimal Jaccard of $0.6$ was used on the ADNI and of $0.5$ on the WDBC dataset. We used maximal $p$-value of $0.01$, minimal support of $10$ and maximal support of $0.8\cdot |E_{train}|$ (these parameter values are commonly used, see \cite{MihelcicRMRF}).  The resulting datasets contained $\approx 915$ entities per fold and $\approx 560$ entities on the WDBC dataset (deviations in numbers of entities between folds are small). 
The minimal Jaccard index $\geq 0.5$ ensures that utilized redescriptions contain meaningful associations, which are later used to describe predictions of the target DLMs. A  minimal Jaccard value of $0.5$ was used on the WDBC dataset to obtain comparable number of redescription candidates as on the ADNI dataset, on which it is possible to obtain a larger number of highly accurate redescriptions. Data views contained a domain data view, $3$ layers of the Multilayer Perceptron ANN (all but the input layer). On the WDBC dataset, we additionally performed the experiment with the predicted label as additional (last) view.  Obtained rules and redescriptions were optimized with Algorithm \ref{alg:RSel} using $\delta = 0.5$ on the ADNI and $\delta = 0.9$ on the WDBC dataset. We used the maximal precision threshold containing enough redescriptions to cover entities from the training set. We used a $5$-fold cross-validation. Architecture of used models is described in the Supplementary document.  

Results are ranked by the average fidelity score obtained on each dataset. We denote results obtained by the proposed approach as \texttt{ExItNeRdoM}$_{R}$ for redescriptions, \texttt{ExItNeRdoM}$_{r}$ for rules and \texttt{ExItNeRdoM}$_{R_{nt}}$, \texttt{ExItNeRdoM}$_{r_{nt}}$ without predicted labels on the WDBC dataset. Fidelity score on the ADNI dataset is significantly improved when the first hidden layer rules are removed from redescriptions, this is denoted \texttt{ExItNeRdoM}$_{R'}$. Results on the ADNI are: 
\begin{lstlisting}
1. ECLAIRE: $0.559 \pm 0.063$, 
2. ExItNeRdoM$_{R'}$: $0.481 \pm 0.093$, 
3. ExItNeRdoM$_{r}$: $0.464 \pm 0.072$, 
4. REM-D: $0.451 \pm 0.036$, 
5. DeepRED_C5:  $0.448 \pm 0.029$, 
6. pedagogical: $0.44 \pm 0.082$, 
7. ExItNeRdoM$_{R}$: $0.431 \pm 0.091$, 
8. REM-T: $0.362 \pm 0.068$
\end{lstlisting}
\noindent Differences between the approaches are not statistically significant.

\noindent Results on the WDBC are: 
\begin{lstlisting}
1. ExItNeRdoM$_{R}$: $0.977 \pm 0.01$, 
2. ExItNeRdoM$_{R_{nt}}$: $0.967 \pm 0.01$, 
3. REM-D: $0.898 \pm 0.021$, 
4. DeepRED_C5: $0.898 \pm 0.021$, 
5. ECLAIRE: $0.896 \pm 0.03$, 
6. pedagogical: $0.896 \pm 0.033$, 
7. ExItNeRdoM$_{r}$: $0.879 \pm 0.03$, 
8. ExItNeRdoM$_{r_{nt}}$: $0.877 \pm 0.04$, 
9. REM-T: $0.868 \pm 0.032$
\end{lstlisting}

\noindent Experimental results demonstrate that redescriptions can be a powerful tool to describe predictions of ANNs (even without knowledge on predicted targets). An important factor might be the increased expressive power obtained by using a more versatile query language (use of conjunction, disjunction, negation). 

Redescriptions describing model predictions on the first fold on the ADNI dataset are available in Table \ref{tab:REx}. Rules describing neuron activations in the first hidden layer ($q4$) are very complex. Dropping these queries yields significantly better results on this dataset.
 
\begin{table}[ht!]
\caption{Redescription examples obtained on the first fold on the ADNI dataset to describe predictions made by the trained Multilayer perceptron neural network model. Final redescription is a tuple of rules ($q_1, q_2, q_3, q_4)$. - denotes that the methodology did not produce a rule for the SMC class. The main reason is it being heavily underrepresented in the dataset. }
\label{tab:REx}
\resizebox{\textwidth}{!}{%
\centering
\begin{tabular}{ |c|c|c|c| } 
\hline
Queries & Redescription & class \\
\hline
$q_1:$& $-3.3 \leq adas_{13} \leq -0.6\ \wedge\ 0.29 \leq apoe_4\leq 4.88\ \wedge\ -2.5 \leq midtemp\leq 2.5$ & AD \\ 
\cdashline{1-2}
$q_2:$& $\neg\ (0.0 \leq n_{AD,3}\leq 0.09)$ &  \\ 
\cdashline{1-2}
$q_3:$& $(-0.99 \leq n_{7,2}\leq 0.58\ \wedge\ -0.76\leq n_{3,2}\leq 0.99\ \wedge -0.25\leq n_{33,2}\leq 0.99\ \wedge\ $&  \\ 
& $ -0.58\leq n_{18,2}\leq 0.99\ \wedge\ -0.99\leq n_{6,2}\leq 0.25) \ \vee\ ( -0.64\leq n_{33,2}\leq 0.19\ \wedge$&  \\ 
& $ -0.46\leq n_{11,2}\leq 0.98\ \wedge\ -0.64\leq n_{18,2}\leq 0.6\ \wedge\ -0.99\leq n_{6,2}\leq -0.72)$&  \\ 
\cdashline{1-2}
$q_4:$&  $(-0.47\leq n_{20,1}\leq 0.99\ \wedge\ -0.29\leq n_{35,1}\leq 0.99\ \wedge\ 0.17\leq n_{3,1}\leq 0.99\ \wedge$ &  \\ 
&  $0.62\leq n_{44,1}\leq 0.99\ \wedge\ -0.69\leq n_{60,1}\leq 0.98\ \wedge\ -0.77\leq n_{31,1}\leq 0.99)\ \vee$ &  \\ 
&  $(0.51\leq n_{3,1}\leq 0.88\ \wedge\ 0.47\leq n_{24,1}\leq 0.99\ \wedge\ -0.32\leq n_{44,1}\leq 0.54\ \wedge$ &  \\ 
&  $-0.43\leq n_{57,1}\leq 0.82\ \wedge\ -0.30\leq n_{4,1}\leq 0.88)\ \vee\ (-0.55\leq n_{4,1}\leq 0.99\ \wedge$ &  \\ 
&  $-0.49\leq n_{1,1}\leq 0.99\ \wedge\ -0.81\leq n_{51,1}\leq 0.94\ \wedge\ -0.58\leq n_{15,1}\leq 0.99\ \wedge$ &  \\ 
&  $0.64\leq n_{44,1}\leq 0.99\ \wedge\ 0.17\leq n_{24,1}\leq 0.99\ \wedge\ -0.99\leq n_{58,1}\leq -0.41)\ \vee$ &  \\
&  $(-0.72\leq n_{49,1}\leq 0.65\ \wedge\ -0.08\leq n_{2,1}\leq 0.47\ \wedge\ -0.25\leq n_{44,1}\leq 0.54\ \wedge$ &  \\
&  $-0.42\leq n_{20,1}\leq 0.83\ \wedge\ -0.91\leq n_{16,1}\leq -0.51\ \wedge\ -0.91\leq n_{30,1}\leq 0.79)$ &  \\\hline
$q_1:$& $-0.58 \leq apoe_4\leq 4.88$ & LMCI \\ \cdashline{1-2}
$q_2:$& $\neg\ (0.16 \leq n_{cn,3}\leq 0.99\ \wedge\ 0.0 \leq n_{emci,3}\leq 0.11)$ &  \\ \cdashline{1-2}
$q_3:$& $-0.99 \leq n_{10,2}\leq 0.03$ &  \\ \cdashline{1-2}
$q_4:$& $-0.40 \leq n_{45,1}\leq 0.99\ \wedge\ -0.99 \leq n_{17,1}\leq 0.84$ &  \\\hline 
$q_1:$& $\neg\ (-3.3 \leq adas_{13}\leq -1.76\ \wedge\ 0.0\leq ecog\leq 3.1)$ & EMCI \\ \cdashline{1-2}
$q_2:$& $0.0 \leq n_{AD,3}\leq 0.11$ &  \\ \cdashline{1-2}
$q_3:$& $-0.99\leq n_{23,2}\leq -0.29 \ \vee\ (\neg\ (-0.50\leq n_{3,2}\leq 0.99\ \wedge\ -0.22\leq n_{23,2}\leq 0.99\ \wedge$ &  \\ 
& $-0.99\leq n_{21,2}\leq -0.69 \ \wedge\ -0.36\leq n_{18,2}\leq 0.99\ \wedge -0.79\leq n_{11,2}\leq 0.99))$ &  \\ \cdashline{1-2}$q_4:$& $\neg\ (0.18\leq n_{3,1}\leq 0.99\ \wedge -0.29\leq n_{35,1}\leq 0.99\ \wedge\ -0.78\leq n_{18,1}\leq 0.95\ \wedge$ &  \\ 
& $-0.77\leq n_{31,1}\leq 0.99\ \wedge -0.48\leq n_{20,1}\leq 0.99\ \wedge\ 0.62\leq n_{44,1}\leq 0.99\ \wedge$ &  \\ 
& $-0.68\leq n_{60,1}\leq 0.98)$ &  \\\hline
& - & SMC \\ \hline
$q_1:$& $-0.86 \leq apoe_4\leq 0.29$ & CN \\ \cdashline{1-2}
$q_2:$& $0.0 \leq n_{cn,3}\leq 0.99$ &  \\ \cdashline{1-2}
$q_3:$& $0.12 \leq n_{7,2}\leq 0.99\ \wedge\ -0.89 \leq n_{17,2}\leq 0.93$ & \\ \cdashline{1-2}
$q_4:$& $\neg\ (0.78 \leq n_{1,1}\leq 0.99)$ &  \\ 
\hline
\end{tabular}}
\end{table}

\subsection{Explaining and relating layers of deep learning models}
\label{sec:expreldl}



We perform multiple experiments relating layers of DLMs with original data, or relating layers of two or more different DLMs. Relating targeted layers of DLM with the domain data fosters local,  global and cohort post hoc explainability \cite{ExpSurv}. It provides information about the activations of individual neurons or groups of neurons for various subsets of data entities, relating them with domain data attribute values, forming equivalence relation between neuronal activations and data attribute values. Information in a text-based format is provided to the domain experts on the role of the described neurons and groups of neurons. Relating layers of different models allows studying various effects of training different architectures of DLMs on the same data or studying the effects of different initializations, additional training or transfer learning.  

\subsubsection{The experimental setup}
The \texttt{ExItNeRdoM} is compared to the ReReMi \cite{GalbrunBW} and the GCLUS-RM \cite{MihelcicRMRF} on data containing maximally two views, and to the GCLUS-RM \cite{MihelcicRMRF} on other data. For each tested approach, we report the number of individually explained neurons and the number of neurons explained in the interactions. These metrics will provide an understanding of the amount of information obtainable from some DLM by the tested approaches. We also report the average Jaccard index of all produced redescriptions and the number of highly accurate redescriptions produced by the approach (accuracy $\geq$ 0.7). We used minimal accuracy of $0.3$, maximal $p$-value of $0.01$, minimal support of $10$ and the maximal support of $0.8\cdot |E|$ in all experiments (the last three parameter values are standardly used, see \cite{MihelcicRMRF}). The minimal Jaccard index of $0.3$ is used to allow describing a larger subset of neurons of DLMs and to assess the extent of the ability of regular redescription mining approaches to be used for this task.  Redescription examples, obtained by the \texttt{ExItNeRdoM}, allowing assessment of the usefulness of the proposed approach for explaining different aspects of the targeted DLMs are provided in Section \ref{ssec:rex}.

We relate penultimate layers of three different CNN networks \cite{mnist} containing $256$ neurons,  trained on the MNIST dataset \cite{mnist}, with $2,\ 3$ and $4$ layers.  $3$ sets of $256$ neurons are related, $768$ neurons in total using $5000$ entities. The corresponding experiment is denoted MNIST$_{CNN_{1,2,3}}$. Next, we relate penultimate layers, containing $512$ neurons each, of $4$ different ResNet networks \cite{ResNet} with $7$ layers, trained with different random initializations on the CIFAR-10 dataset \cite{krizhevsky2009learning}. $4$ sets of $512$ neurons are related, $2048$ neurons in total using $5000$ entities. The corresponding experiment is denoted RESNET$_{1,2,3,4}$.  We relate the penultimate layer, containing $768$ neurons, of a BERT network \cite{BERT} having $14$ layers with the penultimate layer, containing $256$ neurons, of the BertMini network \cite{turc2019well} with $6$ layers, trained on the AGNews dataset \cite{agnews}. $1024$ neurons are related in total using $7600$ entities. The corresponding experiment is denoted AGNews$_{B-Bmini}$. Next, we relate a penultimate layer (with $100$ neurons) of the Multilayer Perceptron with $3$ layers to the original MNIST data (MNIST$_{MLP-Orig}$) using $10000$ entitites and the penultimate layer of the aforementioned BERT model ($768$ neurons in total) with the TFIDF representation of the original textual data (AGNews$_{B-T_{500}}$) using $7060$ entities, where we use only words that occur in at least $500$ texts ($960$ words in total). Finally, we relate a penultimate layer (with $24$ neurons) of the Multilayer Perceptron containing $4$ layers with the features of the original ADNI data (ADNI$_{MLP-Orig}$) using $275$ entities and the penultimate layer (with $64$ neurons) of the Multilayer Perceptron containing $4$ layers with the features of the original Breast Cancer data (WDBC$_{MLP-Orig}$) using $170$ entities. On the last two datasets, we used only the conjunction operators in query construction, which allows obtaining easily interpretable redescriptions, verifiable with the domain data. All redescriptions are created on a set of entities used to test the DLMs. This is done to avoid any biases introduced during the DLM training procedure and to demonstrate the ability of the methodology to be used on novel data (data of interest or studied). It also allows using the data used to train the DLMs as a validation set to evaluate predictive and generalization abilities of the obtained redescriptions. 

\subsubsection{The evaluation results}

\begin{table}[ht!]
\caption{The number of produced redescriptions ($|\mathcal{R}|$), the number of individually described neurons of the DLM (\#$\mathcal{N}_{ind}$), the number of described neurons in interactions of the DLM (\#$\mathcal{N}_{int}$), the average redescription accuracy and standard deviation ($av_J$) and the number of non-redundant accurate redescriptions reported by each approach ($|\{R_{J(R)\geq 0.7}\}|$ ). Mark - in the table denotes that methodology is not applicable to the studied data}
\label{tab:REv}
\begin{center}
\resizebox{\textwidth}{!}{%
\begin{tabular}{ |c|c|c|c|c|c|c| } 
\hline
Experiment & Algorithm & $|\mathcal{R}|$ & \#$\mathcal{N}_{ind}$ & \#$\mathcal{N}_{int}$ & $av_J$ &  $|\{R_{J(R)\geq 0.7}\}|$  \\
\hline
\multirow{3}{7em}{MNIST$_{CNN_{1,2,3}}$} & ExItNeRdoM & $100043$& $\mathbf{681}$ & $\mathbf{768}$ & $0.48\pm 0.16$ & $\mathbf{12332}$ \\ 
& GCLUS-RM & $500$ & $18$ & $496$ & $\mathbf{0.78\pm 0.10}$ & $411$ \\ 
 & ReReMi & - & - & - & - & - \\ 
\hline
\multirow{3}{7em}{RESNET$_{1,2,3,4}$} & ExItNeRdoM & $184380$ & $\mathbf{1438}$ & $\mathbf{2040}$ & $0.39\pm 0.09$& $\mathbf{1901}$ \\ 
 & GCLUS-RM & $300$ & $4$ & $233$ & $\mathbf{0.58 \pm 0.10}$ & $65$ \\ 
 &  ReReMi & - & - & - & - & - \\ \hline
\multirow{3}{7em}{AGNews$_{B - Bmini}$} & ExItNeRdoM & $317284$ & $\mathbf{1020}$ & $\mathbf{1024}$  & $0.78 \pm 0.13$  &   $>\mathbf{20000}$\\ 
& GCLUS-RM & $1000$ & $16$  & $267$  & $0.86\pm 0.10$ & $950$ \\ 
 & ReReMi & $1015$  &  $427$ & $783$  & $\mathbf{0.93\pm 0.13}$  & $968$  \\ 
\hline
\multirow{3}{7em}{MNIST$_{MLP-\text{Orig}}$} & ExItNeRdoM & $314473$ & $\mathbf{99}$  & $\mathbf{100}$  &  $\mathbf{0.78 \pm 0.08}$ &  $>\mathbf{20000}$ \\ 
& GCLUS-RM & $1000$  &  $11$  & $92$  & $0.77 \pm 0.15$ & $830$ \\ 
 & ReReMi & $0$  &  $0$ &  $0$ & $0 \pm 0$  & $0$  \\ 
\hline
\multirow{3}{7em}{AGNews$_{B-T_{500}}$} & ExItNeRdoM & $18235$  & $\mathbf{612}$ & $\mathbf{749}$  & $\mathbf{0.76 \pm 0.10}$  & $\mathbf{15454}$  \\ 
& GCLUS-RM & $604$  & $5$  & $76$  & $0.71 \pm 0.14$ & $428$ \\ 
 & ReReMi &  $0$  &  $0$ &  $0$ & $0 \pm 0$  & $0$  \\ 
\hline
\multirow{3}{7em}{ADNI$_{MLP-\text{Orig}}$} & ExItNeRdoM & $295$ &  $\mathbf{18}$ & $\mathbf{18}$ & $0.66 \pm 0.12$ & $\mathbf{81}$ \\ 
& GCLUS-RM & $41$  &  $4$ & $17$  &  $0.66 \pm 0.12$ & $15$ \\ 
 & ReReMi & $29$  &  $9$ &  $13$ & $\mathbf{0.94\pm 0.04}$  &  $29$  \\ 
\hline
\multirow{3}{7em}{WDBC$_{MLP-\text{Orig}}$} & ExItNeRdoM & $366$ & $\mathbf{51}$ & $\mathbf{52}$  & $0.77 \pm 0.15$  & $\mathbf{184}$  \\ 
& GCLUS-RM & $13$  & $4$ &  $13$  &  $0.83 \pm 0.16$ & $10$ \\ 
 & ReReMi & $34$ &  $9$ & $10$ & $\mathbf{0.99\pm 0.0}$  &  $34$ \\ 
\hline
\end{tabular}}
\end{center}
\end{table}

The evaluation results can be seen in Table \ref{tab:REv}. These results demonstrate that the \texttt{ExItNeRdoM} correctly executes its purpose of describing neurons of targeted neural networks. \texttt{ExItNeRdoM} significantly outperforms existing redescription mining approaches, the \texttt{ReReMi} and the \texttt{GCLUS-RM} with respect to the number of individually described neurons (relating a neuron and its activation value interval with neurons contained in  layers of other DLMs or with attributes of original data) and neurons described in interactions (groups of neurons and their activations are related to groups of neurons contained in layers of other networks or attributes of original data). Since the \texttt{ExItNeRdoM} exhaustively describes all neurons of the targeted layers or the whole DLM, the overall number of produced redescriptions is significantly larger than produced by the redescription mining approaches. However, unlike with regular redescription mining algorithms, obtained redescriptions are grouped per neuron, allowing easy analyses of obtained knowledge. As a consequence of explaining large number of neurons individually and in interactions, the average accuracy of redescriptions produced by the \texttt{ExItNeRdoM} is, in principle, lower than the average accuracy produced by the redescription mining approaches, although there are cases where the \texttt{ExItNeRdoM} even outperforms the competitors. However, the average accuracy of redescriptions produced by the \texttt{ExItNeRdoM} can be further improved by utilizing more computation time and enforcing stricter accuracy threshold. Redescription mining approaches focus on accuracy, thus they describe a much smaller subset of neurons that can be readily described with higher accuracy, mostly in interactions. Since they treat the DLM data as any multi-view dataset, these algorithms lack the knowledge about the neuronal structure of the DLM, and by its heuristic nature fail to describe large portions of available neurons. As the last column of Table \ref{tab:REv} shows, the \texttt{ExItNeRdoM} creates a significantly larger number of accurate redescriptions than competitors. Another downside of regular redescription mining approaches is that they can generate a number of very similar redescriptions, providing very limited information about the DLM. This situation occurred in  redescription sets produced by the \texttt{ReReMi} and the \texttt{GCLUS-RM} approaches on the Breast Cancer dataset.

\subsection{Redescription examples obtained with \texttt{ExItNeRdoM}}
\label{ssec:rex}

We present several redescription examples obtained on the ADNI$_{MLP-\text{Orig}}$, MNIST$_{CNN_{1,2,3}}$, RESNET$_{1,2,3,4}$ and the AGNews$_{B-T_{500}}$ experiment to demonstrate information about the studied DLMs obtainable by the \texttt{ExItNeRdoM}. The corresponding datasets were obtained by computing the DLM activations for entities on the test set and utilizing domain data attributes of the test set. The ability of the \texttt{ExItNeRdoM} to use information in the test set without utilizing target label information is an important advantage compared to the majority of existing XAI methods. The obtained redescriptions are presented in a table format displaying queries and corresponding redescription quality measures, as is usual in the descriptive tasks such as redescription mining, see for example the evaluation in \cite{GalbrunBW}.

Since all redescriptions are created on the data set used to test the DLMs, we evaluate them and compute the corresponding statistics on the set used to train the DLMs. If there is no significant drop in redescription accuracy, when evaluated on the set used to train our DLMs, we conclude that the evaluated redescription captured important associations of neurons of the studied DLM. Using set used to train DLMs to evaluate redescriptions has a further advantage that the available train set target labels can be used as an independent filter to search for interesting redescriptions (these describing a class of interest or having an interesting distribution of target labels).  We extracted two sets of redescriptions: a) redescriptions with accuracy $>0.6$ on the train set with maximal support of $0.5\cdot |E|$ (they are focused), and have Shannon Entropy of target labels on the train set less than half of the maximum measured entropy, $0.5\cdot max_{ShEnt}$, (large majority of re-described entities belongs to one class), b) redescriptions with accuracy $>0.6$ on the train set, have a maximal support of $0.5\cdot |E|$ (they are focused), and have Shannon Entropy of target labels on the train set less than $0.9\cdot max_{ShEnt}$. This allows detection of neurons that are strongly linked to the prediction of a target label, and study of their activations. Selected redescriptions had minimal accuracy of $0.5$ and maximal $p$-value of $0.01$ on the DLMs test data, thus they capture relevant associations.

\subsubsection{Obtained redescriptions}
Table \ref{tab:ADNI} presents selected redescriptions on the ADNI$_{MLP-\text{Orig}}$ dataset along with redescription accuracy on the train and the test set, the value of entropy and the distribution of target labels on the train set. Redescriptions are created using the test set of the DLMs in an unsupervised manner, thus the set used to train the DLMs is used to evaluate obtained redescriptions (often $J_{train}<J_{test}$).

Set $a)$ on the ADNI$_{MLP-\text{Orig}}$ dataset contains $4$ redescriptions and the set b) $38$ redescriptions. Analysing redescription set a) we noticed increased activation of $n_{9,3}$ in a redescription describing cognitively healthy and patients with small cognitive concern, increased activations of $n_{21,3}$ and $n_{2,3}$ in synergy in a redescription describing mostly patients diagnosed with Alzheimer's disease, and some patients in the state of late mild cognitive impairment. Decreased activation of $n_{4,3}$ in synergy with increased activation of $n_{0,3}$ in a redescription describing mostly patients diagnosed with the Alzheimer's disease, similarly to the low activation of $n_{6,3}$ contained in a redescription describing mostly patients diagnosed with the Alzheimer's disease. By observing redescriptions in redescription set b), we also notice that lower activation levels of $n_{21,3}$ ($<1.6$) occur in redescriptions mostly describing cognitively normal people or patients with small cognitive concern or early mild cognitive impairment. A similar observation is true for the $n_{2,3}$. These observations are additionally validated by the values of the indicator called Clinical Dementia Rating Scale – Sum of Boxes (\texttt{CDRSB}, see \cite{o2008staging}) contained in the second query of these redescriptions. 

\begin{table}[ht!]
\caption{Selected redescriptions obtained on the ADNI$_{MLP-\text{Orig}}$ dataset, describing a homogeneous set of entities with respect to target labels from the part of the data used to train the MLP models. CDRSB levels highly correlate with $n_{21,3}$ activations. Redescription support set size on the part of the data used to train the MLP models is denoted $supp(R)_{tr}$ and the part used for testing (used to obtain redescriptions) is denoted $supp(R)_{ts}$. The Distribution$_{tr}$ column denotes the target label distribution of entities in $supp(R)_{tr}$.}
\label{tab:ADNI}
\begin{center}
\resizebox{\textwidth}{!}{%
\begin{tabular}{ |c|c|c|c|c|c| } 
\hline
Redescriptions &  $J_{train}$ & $J_{test}$ & $supp(R)_{tr-ts}$ &  Distribution$_{tr}$  \\
\hline
$q_{1,1}$: $1.32 \leq n_{9,3}\leq 3.3$ &  \multirow{2}{2em}{$0.63$} &  \multirow{2}{2em}{$0.78$} & \multirow{2}{4em}{$122-72$} & \multirow{2}{11em}{AD: 0, LMCI: 0, EMCI: 0, SMC: 28, CN: 94} \\ \cdashline{1-1}
$q_{1,2}$: $0 \leq$\texttt{CDRSB}$\leq 0\ \wedge\ $ $1 \leq$\texttt{EcogSPMem}$\leq 1.6$&  & & & \\
\hline
$q_{2,1}$: $1.4 \leq n_{21,3}\leq 3.1\ \wedge\ 5.25\leq n_{2,3} \leq 13.97$ &  \multirow{2}{2em}{$0.75$}  &  \multirow{2}{2em}{$0.86$} &  \multirow{2}{4em}{$67-24$} &  \multirow{2}{11em}{AD: 63, LMCI: 4, EMCI: 0, SMC: 0, CN: 0}\\\cdashline{1-1}
$q_{2,2}$: $4 \leq$\texttt{CDRSB}$\leq 10$ & & & & \\
\hline
$q_{3,1}$: $0.0 \leq n_{21,3}\leq 1.6\ \wedge\ 0\leq n_{20,3} \leq 0\ \wedge$ &  \multirow{3}{2em}{$0.67$}  &  \multirow{3}{2em}{$0.60$} &  \multirow{3}{4em}{$215-119$} &  \multirow{3}{11em}{AD: 0, LMCI: 7, EMCI: 35, SMC: 62, CN: 111}\\
$1.26 \leq n_{4,3}\leq 3.76\ \wedge\ 0 \leq n_{2,3}\leq 2.7$ & & & & \\\cdashline{1-1}
$q_{3,2}$: $0 \leq$\texttt{CDRSB}$\leq 0.5$ & & & & \\
\hline
$q_{4,1}$: $0 \leq n_{4,3}\leq 0.55\ \wedge\ 1.5\leq n_{0,3} \leq 13.4$ &  \multirow{2}{2em}{$0.66$}  &  \multirow{2}{2em}{$0.74$} &  \multirow{2}{4em}{$117-43$} &  \multirow{2}{11em}{AD: 98, LMCI: 16, EMCI: 3, SMC: 0, CN: 0}\\\cdashline{1-1}
$q_{4,2}$: $2.5 \leq$\texttt{CDRSB}$\leq 10$ & & & & \\
\hline
\end{tabular}}
\end{center}
\end{table}

Set a) on the MNIST$_ {CNN_{1,2,3}}$ dataset contains $11365$ redescriptions and the set b) $24490$ redescriptions. Using filtering with Jaccard index threshold of $0.9$ on set a), we discovered the first redescription presented in Table \ref{tab:MNIST}. It describes neuron activations of three different layers of $3$ different CNN networks that occur almost exclusively on pictures displaying the digit $1$. High accuracy of this redescription indicates that the discovered combinations of neurons of each network discriminate digit $1$ with high accuracy. The second redescription was obtained by searching for redescriptions with homogeneous class distribution in a support set. This redescription contains neuron activations that exclusively occur on pictures containing a digit $3$. 

\begin{table}[ht!]
\caption{Selected redescriptions, obtained on the MNIST$_ {CNN_{1,2,3}}$ dataset, contain neuron activations that co-occur almost exclusively for images depicting digit $1$ (top) and digit $3$ (bottom).  Each query contains a subset of neurons from the penultimate layer of one of the used CNN networks with the corresponding activation intervals. The discovered non-linear relations between subsets are described using conjunction, disjunction and negation logical operators. The first, highly accurate, redescription proves that detected neuron subsets, belonging to $3$ different CNNs, have identical function. Table attributes are as in Table \ref{tab:ADNI}.}
\label{tab:MNIST}
\begin{center}
\resizebox{\textwidth}{!}{%
\begin{tabular}{ |c|c|c|c|c|c| } 
\hline
Redescriptions &  $J_{train}$ & $J_{test}$ & $supp(R)_{tr-ts}$ &  Distribution$_{tr}$  \\
\hline
$q_{1,1}$: $\neg\ (-9.72 \leq CNN_{1}n_{140,1}\leq 0.4)$ &  \multirow{16}{2em}{$0.95$} &  \multirow{16}{2em}{$0.97$} & \multirow{16}{4em}{$6384-1101$} & \multirow{16}{11em}{0: 0, 1: 6382, 2: 0, 3: 0, 4: 0, 5: 0, 6: 0, 7: 0, 8: 2, 9: 0} \\ \cdashline{1-1}
$q_{1,2}$: $-1.69 \leq CNN_{2}n_{203,2}\leq 0.8\ \wedge$  &  & & & \\ 
$-1.67\leq CNN_{2}n_{9,2}\leq 3.55\ \wedge\ $ &  & & &\\
$-1.91\leq CNN_{2}n_{157,2}\leq 0.62\ \wedge\ $ &  & & &\\
$-1.05\leq CNN_{2}n_{181,2}\leq 1.88\ \wedge\ $ &  & & &\\
$0.79\leq CNN_{2}n_{234,2}\leq 2.88\ \wedge\ $ &  & & &\\ \cdashline{1-1}
$q_{1,3}$: $-0.87\leq CNN_{3}n_{233,3}\leq 1.2\ \wedge\ $ &  & & &\\
$-1.39\leq CNN_{3}n_{208,3}\leq 0.17\ \wedge\ $ &  & & &\\
$-2.33\leq CNN_{3}n_{229,3}\leq -1.35\ \wedge\ $ &  & & &\\
$0.77\leq CNN_{3}n_{222,3}\leq 1.96\ \wedge\ $ &  & & &\\
$0.16\leq CNN_{3}n_{116,3}\leq 1.36\ \vee\ $ &  & & &\\
$(0.47\leq CNN_{3}n_{222,3}\leq 2.12\ \wedge\ $ &  & & &\\
$0.25\leq CNN_{3}n_{173,3}\leq 1.81\ \wedge\ $ &  & & &\\
$0.36\leq CNN_{3}n_{123,3}\leq 2.04)\ \vee\ $ &  & & &\\
$(-0.08\leq CNN_{3}n_{130,3}\leq 1.16\ \wedge\ $ &  & & &\\
$0.58\leq CNN_{3}n_{146,3}\leq 1.57) $ &  & & &\\
\hline
$q_{2,1}$: $-5.33\leq CNN_{1}n_{219,1}\leq 0.2\ \wedge$ &  \multirow{10}{2em}{$0.64$}  &  \multirow{10}{2em}{$0.70$} &  \multirow{10}{4em}{$3442-621$} &  \multirow{10}{11em}{0: 0, 1: 0, 2: 0, 3: 3442, 4: 0, 5: 0, 6: 0, 7: 0, 8: 0, 9: 0}\\
$0.83\leq CNN_{1}n_{229,1}\leq 3.1\ \wedge$ & & & & \\
$-2.79\leq CNN_{1}n_{255,1}\leq 0.4\ \wedge$ & & & & \\
$-5.9\leq CNN_{1}n_{231,1}\leq -0.5$ & & & & \\ \cdashline{1-1}
$q_{2,2}$: $\neg\ (-6.23\leq CNN_{2}n_{34,2}\leq 2.03)$ & & & & \\ \cdashline{1-1}
$q_{2,3}$: $2.37\leq CNN_{3}n_{232,3}\leq 4.51\ \vee$ & & & & \\
$(2.41\leq CNN_{3}n_{242,3}\leq 4.0\ \wedge$ & & & & \\
$1.05\leq CNN_{3}n_{234,3}\leq 2.65\ \wedge$ & & & & \\
$1.97\leq CNN_{3}n_{232,3}\leq 2.74\ \wedge$ & & & & \\
$-0.20\leq CNN_{3}n_{139,3}\leq 1.23)$ & & & & \\
\hline
\end{tabular}}
\end{center}
\end{table}

Sets a) and b) on the AGNews$_{B-T_{500}}$ datasets contain $2$ redescriptions presented in Table \ref{tab:AGnews}. Both redescriptions contain neurons and frequencies of words that occur in texts classified in the category \emph{business}. The important attributes from the original data, associated with the discovered neurons are \emph{wednesday} (the day of the week), \emph{target} (can be deadline, goals etc.) and \emph{aspx} (active server pages extended - the web page format of the \texttt{Microsoft ASP.NET}).


\begin{table}[ht!]
\caption{Selected redescriptions, obtained on the AGNews$_{B-T_{500}}$ dataset, contain activations of subsets of neurons of a BERT model $B$ and the corresponding relative occurrence of words that appear together exclusively for texts discussing business. Moreover, these redescriptions explain that the detected activations of the subgroups of neurons prevalently occur with  words \emph{aspx}, \emph{Wednesday} and \emph{target}, which are a reasonable fit for an article discussing business. Table attributes are as in Table \ref{tab:ADNI}.}
\label{tab:AGnews}
\begin{center}
\resizebox{\textwidth}{!}{%
\begin{tabular}{ |c|c|c|c|c|c| } 
\hline
Redescriptions &  $J_{train}$ & $J_{test}$ & $supp(R)_{tr-ts}$ &  Distribution$_{tr}$  \\
\hline
$q_{1,1}$: $0.08 \leq aspx\leq 0.14\ \wedge\ 0 \leq wednesday\leq 0.08$ &  \multirow{8}{2em}{$0.64$} &  \multirow{8}{2em}{$0.88$} & \multirow{8}{4em}{$856-57$} & \multirow{8}{11em}{Sport: 0, Sci/Tech: 0, World: 0, Business: 856} \\
$\ \wedge\ 0.08\leq target\leq 0.13$ &  & & &\\ \cdashline{1-1}
$q_{1,2}$: $-0.73 \leq Bn_{593,12}\leq -0.63\ \wedge$  &  & & & \\
$-0.73\leq Bn_{381,12}\leq -0.66\ \wedge\ $ &  & & &\\
$-0.18\leq Bn_{421,12}\leq -0.07\ \vee\ $ &  & & &\\
$(0.74\leq Bn_{666,12}\leq 0.78\ \wedge\ $ &  & & &\\
$-0.03\leq Bn_{569,12}\leq 0.11\ \wedge\ $ &  & & &\\
$-0.5\leq Bn_{334,12}\leq -0.33)$ &  & & &\\
\hline
$q_{2,1}$: $0.09 \leq target\leq 0.10\ \vee\ (0.08 \leq aspx\leq 0.14\ \wedge\ $ &  \multirow{6}{2em}{$0.67$} &  \multirow{6}{2em}{$0.87$} & \multirow{6}{4em}{$862-59$} & \multirow{6}{11em}{Sport: 0, Sci/Tech: 0, World: 0, Business: 862} \\
$0.08\leq target\leq 0.13) $ &  & & &\\ \cdashline{1-1}
$q_{2,2}$: $0.74 \leq Bn_{666,12}\leq 0.79\ \wedge$  &  & & & \\
$-0.99\leq Bn_{256,12}\leq -0.97\ \wedge\ $ &  & & &\\
$-0.47\leq Bn_{469,12}\leq -0.16\ \vee\ $ &  & & &\\
$-0.51\leq Bn_{14,12}\leq -0.05$ &  & & &\\
\hline
\end{tabular}}
\end{center}
\end{table}

Set a) contains $3337$ and set b) $8869$ redescriptions on the CIFAR-10 \par \noindent (RESNET$_{1,2,3,4}$) dataset. 
$5$ redescriptions from set a) are presented in Table \ref{tab:Resnet}. They describe entities depicting \emph{automobile}, \emph{frog}, \emph{ship}, \emph{truck} and \emph{horse}.

\begin{table}[ht!]
\caption{Selected redescriptions, containing subsets of neurons and their activations from $4$ different ResNet models ($C_1, C_2, C_3, C_4$), trained on the CIFAR-10 dataset (RESNET$_{1,2,3,4}$ in Table \ref{tab:REv}), demonstrate shared function of neuronal subsets on a majority of entities. The fact that they describe a homogeneous set of entities, with respect to target labels, makes them potentially good local predictors. As demonstrated, one neuron in one model can have a similar function as a number of neurons in another.
Table attributes are as in Table \ref{tab:ADNI}.}
\label{tab:Resnet}
\begin{center}
\resizebox{\textwidth}{!}{%
\begin{tabular}{ |c|c|c|c|c|c| } 
\hline
Redescriptions &  $J_{train}$ & $J_{test}$ & $supp(R)_{tr-ts}$ &  Distribution$_{tr}$  \\
\hline
$q_{1,1}$: $0.30 \leq C_{1}n_{24,5}\leq 1.56\ \wedge\ 0.0 \leq C_{1}n_{477,5}\leq 0.78\ \wedge$ &  \multirow{5}{2em}{$0.67$} &  \multirow{5}{2em}{$0.59$} & \multirow{5}{4em}{$3670-762$} & \multirow{5}{11em}{deer: 0, automobile: 3669, horse: 0, frog: 0, truck: 0, bird: 0, cat: 0, airplane: 1, ship: 0, dog: 0} \\
$0.57 \leq C_{1}n_{273,5}\leq 3.18$ &  & & &\\ \cdashline{1-1}
$q_{1,2}$: $\neg\ (0.0 \leq C_{2}n_{396,5}\leq 0.83)$ &  & & &\\ \cdashline{1-1}
$q_{1,3}$: $0.0 \leq C_{3}n_{348,5}\leq 0.61\ \wedge\ 0.52 \leq C_{3}n_{208,5}\leq 2.68$ &  & & &\\ \cdashline{1-1}
$q_{1,4}$: $0.73 \leq C_{4}n_{438,5}\leq 2.51\ \wedge\ 0.33 \leq C_{4}n_{177,5}\leq 1.82$ &  & & &\\
\hline
$q_{2,1}$: $0.06 \leq C_{1}n_{428,5}\leq 1.32\ \wedge\ 0.04 \leq C_{1}n_{107,5}\leq 0.82$ &  \multirow{10}{2em}{$0.63$} &  \multirow{10}{2em}{$0.66$} & \multirow{10}{4em}{$2953-611$} & \multirow{10}{11em}{deer: 0, automobile: 0, horse: 0, frog: 2950, truck: 0, bird: 1, cat: 2, airplane: 0, ship: 0, dog: 0} \\
$\ \wedge\ 0.01 \leq C_{1}n_{10,5}\leq 1.26\ \wedge\ 0.82 \leq C_{1}n_{208,5}\leq 2.35\ \wedge$ &  & & &\\
$0.0 \leq C_{1}n_{299,5}\leq 0.18$ &  & & &\\ \cdashline{1-1}
$q_{2,2}$: $\neg\ (0.0 \leq C_{2}n_{82,5}\leq 0.99)$ &  & & &\\ \cdashline{1-1}
$q_{2,3}$: $0.99 \leq C_{3}n_{123,5}\leq 3.23$ &  & & &\\ \cdashline{1-1}
$q_{2,4}$: $0.75 \leq C_{4}n_{42,5}\leq 2.20\ \wedge\ 0.78 \leq C_{4}n_{118,5}\leq 1.96\ \wedge$ &  & & &\\
$0.0 \leq C_{4}n_{448,5}\leq 0.22\ \wedge\ 0.0 \leq C_{4}n_{401,5}\leq 0.21\ \wedge$ &  & & &\\
$0.66 \leq C_{4}n_{491,5}\leq 1.87\ \vee\ (0.0 \leq C_{4}n_{240,5}\leq 0.25\ \wedge$ &  & & &\\
$0.74 \leq C_{4}n_{42,5}\leq 2.20\ \wedge\ 0.0 \leq C_{4}n_{401,5}\leq 0.23\ \wedge$ &  & & &\\
$0.31 \leq C_{4}n_{491,5}\leq 1.87)$ &  & & &\\
\hline
$q_{3,1}$: $0.77 \leq C_{1}n_{380,5}\leq 2.50$ &  \multirow{6}{2em}{$0.63$} &  \multirow{6}{2em}{$0.62$} & \multirow{6}{4em}{$4119-817$} & \multirow{6}{11em}{deer: 0, automobile: 0, horse: 0, frog: 0, truck: 2, bird: 3, cat: 0, airplane: 10, ship: 4103, dog: 1} \\ \cdashline{1-1}
$q_{3,2}$: $ 0.45 \leq C_{2}n_{47,5}\leq 2.43$ &  & & &\\ \cdashline{1-1}
$q_{3,3}$: $\neg\ (0.0 \leq C_{3}n_{85,5}\leq 0.70)$ &  & & &\\ \cdashline{1-1}
$q_{3,4}$: $0.82 \leq C_{4}n_{110,5}\leq 3.15\ \wedge\ 0.73 \leq C_{4}n_{404,5}\leq 2.54$ &  & & &\\
$\ \vee\ (0.65 \leq C_{4}n_{346,5}\leq 1.23\ \wedge\ 0.81 \leq C_{4}n_{110,5}\leq 1.60\ \wedge$ &  & & &\\
$0.15 \leq C_{4}n_{404,5}\leq 0.73)$ &  & & &\\
\hline
$q_{4,1}$: $0.45 \leq C_{1}n_{24,5}\leq 1.73\ \wedge\ 0.60 \leq C_{1}n_{477,5}\leq 2.55$ &  \multirow{6}{2em}{$0.60$} &  \multirow{5}{2em}{$0.63$} & \multirow{5}{4em}{$3240-664$} & \multirow{5}{11em}{deer: 0, automobile: 2, horse: 0, frog: 0, truck: 3237, bird: 0, cat: 0, airplane: 0, ship: 1, dog: 0} \\ \cdashline{1-1}
$q_{4,2}$: $\neg\ (0.0 \leq C_{2}n_{45,5}\leq 0.86)$ &  & & &\\ \cdashline{1-1}
$q_{4,3}$: $1.29 \leq C_{3}n_{322,5}\leq 2.81\ \wedge\ 0.0 \leq C_{3}n_{42,5}\leq 0.33$ &  & & &\\ \cdashline{1-1}
$q_{4,4}$: $1.29 \leq C_{4}n_{410,5}\leq 2.82\ \vee\ (0.10 \leq C_{4}n_{410,5}\leq 1.28$ &  & & &\\
$\ \wedge\ 0.0 \leq C_{4}n_{430,5}\leq 0.52\ \wedge\ 1.26 \leq C_{4}n_{338,5}\leq 1.78)$ &  & & &\\
\hline
$q_{5,1}$: $0.38 \leq C_{1}n_{318,5}\leq 2.43\ \wedge\ 0.66 \leq C_{1}n_{115,5}\leq 2.08$ &  \multirow{6}{2em}{$0.63$} &  \multirow{6}{2em}{$0.64$} & \multirow{6}{4em}{$3529-681$} & \multirow{6}{11em}{deer: 28, automobile: 0, horse: 3493, frog: 0, truck: 0, bird: 0, cat: 0, airplane: 0, ship: 0, dog: 8} \\
$\wedge\ 0.18 \leq C_{1}n_{190,5}\leq 3.15$ &  & & &\\ \cdashline{1-1}
$q_{5,2}$: $\neg\ (0.0 \leq C_{2}n_{166,5}\leq 0.96)$ &  & & &\\ \cdashline{1-1}
$q_{5,3}$: $ 0.79 \leq C_{3}n_{228,5}\leq 2.79$ &  & & &\\ \cdashline{1-1}
$q_{5,4}$: $1.09 \leq C_{4}n_{86,5}\leq 2.52\ \wedge\ 0.84 \leq C_{4}n_{431,5}\leq 3.02$ &  & & &\\
$\ \vee\ (0.29 \leq C_{4}n_{369,5}\leq 2.63\ \wedge\ 0.87 \leq C_{4}n_{268,5}\leq 3.08)$ &  & & &\\
\hline
\end{tabular}}
\end{center}
\end{table}

The accuracy levels, the target label distribution and the original data attribute intervals and meaning demonstrate that the methodology detects neuronal activations of individual neurons or groups of neurons that represent semantically meaningful knowledge, relevant to the domain experts. Moreover, due to the property of redescriptions to capture valid equivalence relations, quantified by the redescription accuracy, they provide information about the functioning of various parts of the used DLM. That is, we obtain a post-hoc explanation of neurons and their activations. For example, relatively high accuracy of presented redescriptions from Table \ref{tab:ADNI} denotes that the neuronal activations (as described) occur in over $60\%$ to close to $80\%$ of the entities if and only if the corresponding Alzheimer's disease indicators are in the predefined intervals, which provides a solid explanation of the function of the chosen neurons with a presented activation intervals.  Similar is true of redescriptions presented in Table \ref{tab:AGnews}, where neuronal activations are associated to business terms \emph{target}, \emph{wednesday} and \emph{aspx}. Redescriptions in Table \ref{tab:MNIST} demonstrate which groups of neurons, in CNNs with different architecture, are related to recognition of various digits. Here, it is visible that one neuron in one DLM can have a very similar, or almost identical (over $95\%$), function as a group of neurons in the other DLM. This can be also seen from examples presented in Table \ref{tab:Resnet}. They demonstrate groups of neurons of the ResNet models, trained with different initializations, related to recognition of various objects. 

Obtained redescriptions can be analyzed further by examining the support sets of redescription queries. This allows determining the degree to which detected neuronal activations are related to the target concepts, and if they have multiple roles and this is only the predominant one. Further, one can analyze distributions of target labels of described entities by each individual neuron (given the activation interval described in the rule). Dependent on the neural network and the application domain, it can happen that a subset of neurons is directly tasked with predicting some selected target label (e.g. $CNN_1n_{140,1}$ in Table \ref{tab:MNIST}), while other neurons appearing in redescription query describe entities from two or more classes, but help in distinguishing the goal class from other classes (given the computed activation interval). One other common pattern is to have multiple neurons describing entities contained in two or more classes, but when they all activate with a predefined activation value, these neurons (jointly) mostly describe entities from only one class group (e.g. query $q_{1,2}$ from Table \ref{tab:MNIST}). Presented redescriptions list all entities from their support set, and all entities from the union of their query supports, thus one can perform \emph{by example} analyses and find counterexamples, entities for which further training or more elaborate data are required to improve the DLM. Discovered redescriptions and queries can potentially be used as novel features  \cite{MihelcicFeatures}, or as local classifiers.  

\section{Discussion}
\label{sec:discuss}
The \texttt{ExItNeRdoM} creates in total a much larger number of redescriptions than comparable redescription mining approaches, however these redescriptions are organized in a number of smaller redescription sets. Two redescription sets are created per neuron. The first set contains redescriptions where one query only has a targeted neuron and its corresponding activation interval, providing explicit information about the function of this neuron in the targeted network, and the second set with one query containing the same neuron in interactions with other neurons of the same layer of the DLM. The second set will often contain information about more complex objects, revealing complex associations required to learn some concept. The interested user can perform very selective analyses. Although understanding individual neurons can often be non sensible to humans \cite{BodriaGGNPR23}, the advantage of the proposed framework is that it can relate them to easily understandable attributes from the domain data. Thus, their roles individually and in a group depicting some complex object can be understood much quicker and easier. 
Redescription sets can be further studied by computing association counts, correlations and support set homogeneity, see \cite{MihelcicADNI}, or can be interactively studied individually or jointly using the redescription set exploration tool InterSet \cite{MihelcicInterSetML}. These properties of the \texttt{ExItNeRdoM} tackle local explainability of the targeted DLM \cite{ExpSurv}. Global analyses can be obtained by targeting all (essential) layers of the DLM. 
Globally, the model can be either analyzed using the InterSet \cite{MihelcicInterSetML}, or a representative subset of redescriptions can be created using procedures such as filtering or redescription set construction  \cite{MihelcicFramework}. The most suitable method of analyses, except neuron activation correlations and relating to original data attributes, would be to analyse model properties by example (locally). One could compute neuron activations for examples of interest, observe their properties described in the original data, and discover a set of redescriptions that describe them. One could also easily discover all redescriptions that indicate discrepancy between computed activations of selected examples and the properties described by the original data, allowing further DLM adaptations. 

The proposed approach requires that DLM representations do not vary in size between data samples. However, with minor adjustment, even such models (e.g. fully convolutional networks \cite{long2015fully}) can be analyzed. To obtain a fixed-size layer representations, one can resize the input samples to equal size or group the samples into subgroups based on size. Analyzing penultimate layers of DLMs performing segmentation, such as the U-net \cite{Unet}, would require significantly larger time compared to the majority of other DLMs of similar size (performing e.g. classification, clustering, encoding etc.) due to high number of neurons in the penultimate layer. Generative and language models can be studied with expectedly high run times, but comparable to their training time. The advantage of the approach is that even for extremely large language models, one can choose a subset of neurons of interest and relate them quickly to domain data or neurons of some other model (e.g. check which neurons in a small model acquired the function and contain information stored in targeted neurons after distillation or similar procedures). 

Many produced redescriptions, by all studied approaches, contain one or more neurons with very wide activation intervals in their queries. Such general information is less useful to understand the role of this neuron in the network. \texttt{ExItNeRdoM} produced redescription sets that describe individual neurons and their activation intervals, obtained using equal-width binning approach by Freedman–Diaconis, provide more targeted information. Obtaining descriptions of individual neurons can be challenging for DLMs trained on the image data, where many shapes and patterns occur in very small number of images and where often some shape or a pattern is detected only by interactions of neurons. However, such interactions are also provided by the approach in a separate set. 

Obtaining information of described detail about DLMs, in an easily understandable textual form, requires internally performing multiple runs of constraint-based redescription mining. Given that these describe only a specific neuron and its interactions, required computations per run can be  performed significantly faster than the execution of an approach for general redescription mining. Many neurons need to be described, but this can be done independently (parallelized). Thus,  utilizing a computationally stronger server is advised. 

Maximal duration of an experiment for \texttt{ExItNeRdoM} was $14$ days (on the multi-view CIFAR-10 dataset) using $30$ threads on a server with \texttt{AMD Opteron 6380}, $48$ cores. The data contains $4$-views, $5000$ entities (equal to the largest in Section \ref{sec:analEnt}), $512$ neurons per network, which amounts to $2048$ in total (over $4x$ more than the largest in Section \ref{sec:analAtt}), and \texttt{WorkSetSize} $=500$ and \texttt{MaxRSSize} $= 1500$ (similar to the smallest in Section \ref{sec:analViews}). Large number of rule candidates causes multiple repetitions of PCT and forest learning and results in many redescription candidates for some neurons. This increases constants from the complexity analyses.  Descriptions of the majority of neurons are obtained in much shorter time, a few days at most. Taking all variables into account, the experiment adheres to the behavior established in Section \ref{sec:scalStudy}.  The longest redescription mining experiments were computed in $16$ days (the \texttt{GCLUS-RM} on multi-view CIFAR-10 data containing $4$-views with equal candidate set parameters as the \texttt{ExItNeRdoM}) and $7$ days (two-view AGNews data with the \texttt{ReReMi}) using a single thread for computation. It should be noted that the \texttt{ReReMi} approach does not support multi-view redescription mining. Execution times of the \texttt{ExItNeRdoM} can be further reduced by decreasing the depth of PCT models and forests and reducing the number of random initializations used. This can reduce execution time by at least five times. 

\section{Conclusions}
\label{sec:conclusion}
A tool \texttt{ExItNeRdoM} is capable of relating neurons of one or more layers of one or more different DLMs. The output is a set of multi-view redescription sets, where each redescription is a tuple of logical formulae. This is a human-understandable representation of related groups of neurons. Neurons can also be related to domain and supplementary data (e.g. target labels, knowledge bases) which provide additional information. The performed experiments show significant correlation between the number of described neurons by the approach, and the similarity between used representations. The \texttt{ExItNeRdoM} has a comparable performance to the state of the art rule extraction approaches, and can be used to explain individual neurons and their interactions across layers and DLMs. The proposed approach significantly outperforms state of the art redescription mining approaches in the coverage of described neurons, is more general (with respect to utilized type of knowledge, architecture and the number of DLMs that can be analyzed). These advantages extend to the existing approach \cite{RmForDL}, that utilizes the ReReMi approach as a black box component. 

\subsection*{Acknowledgement}
\footnotesize{
\noindent The authors acknowledge the assistance in conceptualization, investigation and supervision provided by our esteemed colleague Dr. Tomislav Lipić. Dr. Lipić, unfortunately, left us too early due to critical illness. This work was supported by the DATACROSS (KK.01.1.1.01.0009) and by the Croatian Science Foundation project AIGEN (PZS-2019-02-8525). The ADNI dataset was obtained from the Alzheimer’s Disease Neuroimaging Initiative (ADNI) database.
}

\bibliographystyle{acm}
\bibliography{mybibfile}

\end{document}


\begin{frontmatter}

\title{Supplementary document to: A redescription mining framework for post-hoc explaining and relating deep learning models}
    
\author[myfootnote1]{Matej Mihelčić\fnref{myfootnote}\corref{cor1}}
\ead{matmih@math.hr}
\cortext[cor1]{Corresponding author. E-mail address: matmih@math.hr}
\author[myfootnote2]{Ivan Grubišić\fnref{myfootnote}}
\ead{ivan.grubisic@irb.hr}
\author[myfootnote2]{Miha Keber\fnref{myfootnote}}
\ead{miha.keber@irb.hr}
\address[myfootnote1]{Department of Mathematics, Faculty of Science, University of Zagreb, Bijenička cesta 30, 10000 Zagreb, Croatia}
\address[myfootnote2]{Faculty of Electrical Engineering and Computing, University of Zagreb, Unska cesta 3, 10000 Zagreb, Croatia}
\address[myfootnote]{Ruđer Bošković Institute, Bijenička cesta 54, 10000 Zagreb, Croatia}

\end{frontmatter}

\section{Experiments Conducted in the Manuscript}
\begin{figure}[ht!]
    \begin{center}
    \includegraphics[width=0.8\textwidth]{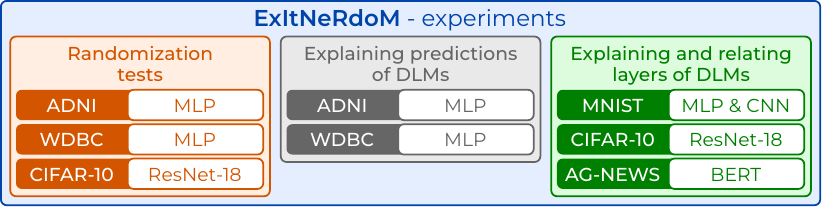}
    \end{center}
    \caption{Datasets and DLM architectures for conducted experiments to validate effectiveness of \texttt{ExItNeRdoM} methodology.}
     \label{fig:exps}
    \end{figure}

    The supplementary material contains glossary with abbreviations, symbols, and methods used in the paper; summary of the deep learning models (DLMs) used in the scalability study experiments; summary of the DLMs used in the randomization test experiments; summary of the DLMs used in the rule extraction experiments; and summary of the DLMs used in the explaining and relating layers of DLMs experiments. All of DLM architectures and datasets for conducted experiments are depicted in Figure \ref{fig:exps}.
    
\section{Glossary}

\begin{table}[!htb]
    \centering
    \scriptsize
    \caption{List of abbreviations used in the paper.}
    \begin{tabular}{l|l}
    AI                                   & Artificial intelligence                                                                                                                                                                                                                                                                                                                                                   \\
    SP                                   & Spearman correlation                                                                                                                                                                                                                                                                                                                                                      \\
    XAI                                  & Explainable AI                                                                                                                                                                                                                                                                                                                                                            \\
    CKA                                  & Central kernel alignment                                                                                                                                                                                                                                                                                                                                                  \\
    SVCCA                                & Singular value canonical correlation analysis                                                                                                                                                                                                                                                                                                                             \\
    AGNews                               & Text classification dataset \citep{agnews}                                                                                                                                                                                                                                                                                                                                                            \\
    ADNI                                 & Alzheimer’s Disease Neuroimaging Initiative                                                                                                                                                                                                                                                                                                                               \\
    WDBC                                 & Breast Cancer Wisconsin (Diagnostic) Data Set                                                                                                                                                                                                                                                                                                                             \\
    AD                                   & Alzheimers disease                                                                                                                                                                                                                                                                                                                                                        \\
    TFIDF                                & Term frequency inverse document frequency                                                                                                                                                                                                                                                                                                                                 \\
    DLM                                  & Deep learning model                                                                                                                                                                                                                                                                                                                                                       \\
    ANN                                  & Artificial neural network                                                                                                                                                                                                                                                                                                                                                 \\
    MLP                                  & Multy-layer perceptron                                                                                                                                                                                                                                                                                                                                                    \\
    CNN                                  & Convolutional neural network                                                                                                                                                                                                                                                                                                                                              \\
    BERT                                 & Transformer DLM architecture \citep{BERT}                                                                                                                                                                                                                                                                                                                                                              \\
    ResNet-18                            & CNN architecture \citep{ResNet}                                                                                                                                                                                                                                                                                                                                                            \\
    GCLUS-RM        & generalized CLUS-RM redescription mining algorithm \citep{MihelcicRMRF}                                                                                                                                                                                                                                                                                                             \\
    ReReMi                               & Redescription mining methodolgy \citep{GalbrunBW}                                                                                                                                                                                                                                                                                                                                                         \\
    PCT                                  & Predictive clustering trees                                                                                                                                                                                                                                                                                                                                               \\
    RF                                   & Random forest.                                                                                                                                                                                                                                                                                                                                                            \\
    \texttt{Eclair}                      & Rule extraction methodology for DLMs \citep{zarlenga2021efficient}                                                                                                                                                                                                                                                                                                                                             \\
    \texttt{Rem-T}                       & Rule extraction methodology for DLMs \citep{shams2021rem}                                                                                                                                                                                                                                                                                                                                                      \\
    \texttt{Rem-D}                       & Rule extraction methodology for DLMs \citep{shams2021rem}                                                                                                                                                                                                                                                                                                                                                      \\
    \texttt{DeepRED}                     & Rule extraction methodology for DLMs \citep{zilke2016deepred}                                                                                                                                                                                                                                                                                                                                                  \\
    texttt{PedC5.0}                      & Rule extraction methodology for DLMs \citep{kola}                                                                                                                          \\
    \texttt{ExItNeRdoM}                  & \begin{tabular}[t]{@{}l@{}}Explaining and interpreting (deep) neural models using \\ redescription mining.\end{tabular}                                      \\
    \end{tabular}
    \label{tab:abbrevations}
    \end{table}

    \begin{table}[!htb]
    \centering
    \scriptsize
    \caption{Symbols used in paper.}
    \begin{tabular}{c|c}
    \textbf{Symbol}                                 & \textbf{Description}                                                                                                                \\ \hline
    $\mathcal{C}$                                   & Deep learning model type.                                                                                                           \\
    $\mathcal{M}$                                   & Deep learning model.                                                                                                                \\
    $\mathcal{P}$                                   & Parameters of a deep learning model.                                                                                                \\
    $\mathcal{N}_{ind}$                             & Number of individually described neurons of the DLM.                                                                                \\
    $\mathcal{N}_{int}$                             & Number of described neurons in interactions of the DLM.                                                                             \\
    $\mathcal{D}$                                   & Tabular dataset, the whole dataset.                                                                                                 \\
    $E$                                             & Entities or samples within the dataset.                                                                                             \\
    $e_k$                                           & $k$-th entity from the dataset.                                                                                                     \\
    $\mathcal{A}$                                   & Set of attributes in the dataset.                                                                                                   \\
    $A$                                             & Attribute from the $\mathcal{A}$.                                                                                                   \\
    $V_i$                                           & Attributes of a view $W_i$.                                                                                                         \\
    $v_{A, min}, v_{A, max}$                        & Minimum and maximum values of attribute $A$.                                                                                        \\
    $\mathcal{W}$                                   & Set of views.                                                                                                                       \\
    $G$                                             & Subset of selected views of interest                                                                                                \\
    $W_i$                                           & Views of the tabular dataset.                                                                                                       \\
    $M$                                             & Number of attributes in a view.                                                                                                     \\
    $N$                                             & Number of output dimension, e.g. number of logits.                                                                                  \\
    $\mathcal{Y}$                                   & Target labels set.                                                                                                                  \\
    $Y_1$                                           & Output classes.                                                                                                                     \\
    $(y_1,...,y_c)$                                 & Target label classes.                                                                                                               \\
    $|\cdot|$                                       & Cardinality of a set.                                                                                                               \\
    $L_k$                                           & Layer $k$ of a deep neural network model.                                                                                           \\
    $L_k^{\mathcal{M}_{\mathcal{C}}^{\mathcal{P}}}$ & Layer $k$ of a model $\mathcal{M}_{\mathcal{C}}^{\mathcal{P}}$.                                                                     \\
    $n_{i,k}$                                       & Neuron $i$ of the layer $k$.                                                                                                        \\
    $id$                                            & Neuron identity.                                                                                                                    \\
    $p_1$                                           & Marginal probabilities.                                                                                                             \\
    RS $\mathcal{F}$                                & Redescription sets family.                                                                                                          \\
    $\mathcal{R}$ or RS                             & Resescription set.                                                                                                                  \\
    $R = (q_1, q_2, \dots, q_n)$                    & Redescription, a tuple of rules.                                                                                                     \\
    $p(R)$                                          & Statistical significance of a redescription.                                                                                                     \\
    $q_i$                                           & Query, rule describing a subset of entities.                                                                                        \\
    $supp(q_i)$                                     & Support set of the query. Set of entities described by a query.                                                                    \\
    $r_i$                                           & \begin{tabular}[t]{@{}c@{}}Class $i$ from PCT algorithm, rule fragment, or one \\ bin from binning procedure.\end{tabular}          \\
    $nt$                                            & Number of trees in PCT RF algorithm.                                                                                                \\
    $av_J$                                          & The average redescription accuracy.                                                                                                                 \\
    $J(R)$                                          & Jaccard index of a redescription $R$.                                                                                                  \\
    $R^2$                                           & The coefficient of determination, i.e. R2 score.                                                                                    \\
    \end{tabular}
    \label{tab:abbrevations}
    \end{table}

    \begin{table}[]
    \centering
    \scriptsize
    \caption{List of methods used in workflow of our algorithms.}
    \begin{tabular}{l|l}
    Abbrevations/Functions               & Description                                                                                                                                                                                                                                                                                                                                                               \\ \hline
    \texttt{WorkSetSize}                 & Defines the number of top redescrition candidates stored during execution of the approach.                                                                                                                                                                                                                              \\
    \texttt{MaxRSSize}                   & Defines the maximal number of redescriptions that can be kept in memory.                                                                                                                                                                                                                                                                                                  \\
    $view(n)$                            & View containing neuron $n$.                                                                                                                                                                                                                                                                                                    \\
    \texttt{completeReds}                & Method for completing incomplete redescriptions. \\
    \texttt{createReds}                  & Method for creating 2-view redescriptions from sets of rules.  \\
    \texttt{transformToRules}($\texttt{PCT}$ | $\texttt{RF}_{\texttt{PCT}}$) & Method for transforming $\texttt{PCT}$ or $\texttt{RF}_{\texttt{PCT}}$ models to rules.                                                                                                                                                                                                                                                                                                                      \\
    \texttt{performBinning}              & Method for uniform attribute binning with a width determined by Freedman-Diaconis rule.                                                                                                                                                                                                                                                                                              \\
    \texttt{createRules}              &  Method to create initial rules from bins. \\
    \texttt{refine}                   &  Method for refining redescriptions by combining neighbouring bins. \\
     \texttt{unifDiv}                     &     Method that uniformly devides neurons that need to be described to the working threads.                                    \\
     \texttt{NcovEntReds}                 &    Number of entities to be covered by redescriptions, for each class value.          \\
     \texttt{NcovEntRules}                &     Number of entities to be covered by rules, for each class value.                                                      \\
     \texttt{chosenRules}                 &    Currently selected set of rules for each class value.                                                                   \\
     \texttt{chosenReds}                  &     Currently selected set of redescriptions for each class value.                                                        \\
     \texttt{classValue}                  &     A set containing all possible class values.                                                                \\
    \texttt{numCov}                      &     Number of entities covered by some rule/redescription.                                                                    \\
     \texttt{equalSupp}                   &    Method that tests equality of support sets.                                                                            \\
     \texttt{countCovered($R$)}           &   Method   that counts the number of entities described by redescription $R$.                  \\
        \end{tabular}
    \label{tab:abbrevations}
    \end{table}

\section{Scalability study of the ExitNeRdoM (Section 5.2.)}

    \begin{table}[!htb]
    \centering
    \scriptsize
    \caption{Test performance of the MLP models used in the \emph{Scalability with respect to the number of entities} (Section 5.2.1.) study. MLP models are initialized with random seeds 313 and 314, trained from scratch on the MNIST train dataset. Each model has an architecture consisting of 784 input features, i.e. image of 28x28 pixels; three hidden layers with 250, 100, and 30 neurons, respectively; and 10 output neurons. Neuron activations were extracted from the last hidden layer, i.e. penultimate layer with 30 neurons, on different number of test set entities, i.e. samples, and used as views.}
    \begin{tabular}{c|c}
    Model        & Test Accuracy \\ \hline
    mlp-p30-s313 & 0.978 \\
    mlp-p30-s314 & 0.979 \\
    \end{tabular}
    \label{tab:scalability-entity}
    \end{table}

    \begin{table}[!htb]
    \centering
    \scriptsize
    \caption{Test performance of the MLP models used in the \emph{Scalability with respect to the number of neurons} (Section 5.2.2.) study. MLP models are initialized with random seeds 411 and 412, trained from scratch on the MNIST train dataset. Each model has an architecture consisting of 784 input features, i.e. image of 28x28 pixels; two hidden layers with 512 and 250 neurons, respectively; and 10 output neurons. Neuron activations of 500 test set samples, i.e. entities, were extracted from the last hidden layer, i.e. penultimate layer with 250 neurons. For neuron scalability study the number of neurons $k$ varies between 50 and 250 with an increment step of 50. Selecting first $k$ neurons of the total neurons for the views. Within the study, to create inter-model redescriptions between 2 MLP models, total number of neurons varies between 100 and 500.}
    \begin{tabular}{c|c}
    Model        & Test Accuracy \\ \hline
    mlp-s411     & 0.980 \\
    mlp-s412     & 0.980 \\
    \end{tabular}
    \label{tab:scalability-entity}
    \end{table}

    \begin{table}[!htb]
    \centering
    \scriptsize
    \caption{Test performance of the MLP models used in the \emph{Scalability with respect to the number of network layers} (Section 5.2.3.) study. MLP models are initialized with different random seeds (311-314) and trained from scratch on the MNIST train dataset. Each model has an architecture consisting of 784 input features, i.e. image of 28x28 pixels; three hidden layers with 250, 100, and 30 neurons, respectively; and 10 output neurons. Neuron activations of 500 test samples, i.e. entities, were extracted from the last hidden layer, i.e. penultimate layer with 30 neurons, and used as views.}
    \begin{tabular}{c|c}
    Model        & Test Accuracy \\ \hline
    mlp-p30-s311 & 0.977 \\
    mlp-p30-s312 & 0.979 \\
    mlp-p30-s313 & 0.978 \\
    mlp-p30-s314 & 0.979 \\
    \end{tabular}
    \label{tab:scalability-entity}
    \end{table}
    
\section{Randomization test experiments (Section 6.1.)}

    \begin{table}[!htb]
    \centering
    \scriptsize
    \caption{Test performance of the models used in the randomization test experiment, i.e. the inter-model analysis of MLP models on the ADNI dataset. Each of 20 MLP models is initialized with different random seeds (211-230) and trained from scratch on the ADNI train dataset. Test accuracy of non-trained models, i.e. models with random weights, is shown in the \emph{Accuracy (random model)} column. Test accuracy of these same models after training is shown in the \emph{Accuracy (trained model)} column. Each model has an architecture consisting of 33 input features; two hidden layers with 64 and 24 neurons, respectively; and 5 output neurons. Neuron activations of 275 test samples extracted from the last hidden layer, i.e. penultimate layer with 24 neurons, are used as views. The effectivness of \texttt{ExItNeRdoM} methodology analysis is performed on redescriptions obtained from pairs of these views.}
    \begin{tabular}{c|c|c}
    Model     & Accuracy (random model) & Accuracy (trained model) \\ \hline
    mlp-s-211 & 0.138           & 0.672            \\
    mlp-s-212 & 0.210           & 0.698            \\
    mlp-s-213 & 0.109           & 0.669            \\
    mlp-s-214 & 0.327           & 0.694            \\
    mlp-s-215 & 0.181           & 0.683            \\
    mlp-s-216 & 0.192           & 0.680            \\
    mlp-s-217 & 0.334           & 0.694            \\
    mlp-s-218 & 0.178           & 0.661            \\
    mlp-s-219 & 0.178           & 0.669            \\
    mlp-s-220 & 0.261           & 0.705            \\
    mlp-s-221 & 0.167           & 0.690            \\
    mlp-s-222 & 0.130           & 0.687            \\
    mlp-s-223 & 0.200           & 0.694            \\
    mlp-s-224 & 0.214           & 0.690            \\
    mlp-s-225 & 0.105           & 0.705            \\
    mlp-s-226 & 0.200           & 0.694            \\
    mlp-s-227 & 0.341           & 0.669            \\
    mlp-s-228 & 0.185           & 0.665            \\
    mlp-s-229 & 0.101           & 0.687            \\
    mlp-s-230 & 0.167           & 0.676           
    \end{tabular}
    \label{tab:rand-adnu}
    \end{table}

    \begin{table}[!htb]
    \centering
    \scriptsize
    \caption{Test performance of the models used in the randomization test experiment, i.e. the inter-model analysis of MLP models on the WDBC dataset. Each of 20 MLP models is initialized with different random seeds (211-230) and trained from scratch on the WDBC train dataset. Test accuracy of non-trained models, i.e. models with random weights, is shown in the \emph{Accuracy (random model)} column. Test accuracy of these same models after training is shown in the \emph{Accuracy (trained model)} column. The models follow a consistent architecture with 30 input features; four hidden layers with 512, 256, 128, and 64 neurons, respectively; and 1 output neuron. Neuron activations of 171 test samples extracted from the last hidden layer, i.e. penultimate layer with 64 neurons, are used as views. The effectivness of \texttt{ExItNeRdoM} methodology analysis is performed on redescriptions obtained from pairs of these views.}
    \begin{tabular}{c|c|c}
    Model     & Accuracy (random model) & Accuracy (trained model) \\ \hline
    mlp-s-211 & 0.257           & 0.941            \\
    mlp-s-212 & 0.385           & 0.941            \\
    mlp-s-213 & 0.614           & 0.964            \\
    mlp-s-214 & 0.391           & 0.947            \\
    mlp-s-215 & 0.614           & 0.953            \\
    mlp-s-216 & 0.362           & 0.953            \\
    mlp-s-217 & 0.602           & 0.953            \\
    mlp-s-218 & 0.485           & 0.947            \\
    mlp-s-219 & 0.385           & 0.941            \\
    mlp-s-220 & 0.298           & 0.953            \\
    mlp-s-221 & 0.385           & 0.964            \\
    mlp-s-222 & 0.614           & 0.941            \\
    mlp-s-223 & 0.385           & 0.953            \\
    mlp-s-224 & 0.614           & 0.953            \\
    mlp-s-225 & 0.385           & 0.959            \\
    mlp-s-226 & 0.608           & 0.959            \\
    mlp-s-227 & 0.397           & 0.959            \\
    mlp-s-228 & 0.385           & 0.947            \\
    mlp-s-229 & 0.315           & 0.953            \\
    mlp-s-230 & 0.614           & 0.953           
    \end{tabular}
    \label{tab:rand-wdbc}
    \end{table}

    \begin{table}[!htb]
    \centering
    \scriptsize
    \caption{Test performance of the models used in the randomization test experiment, i.e. the inter-model analysis of ResNet-18 models on the CIFAR-10 dataset. Each of 20 ResNet-18 models is initialized with different random seeds (211-230) and trained from scratch on the CIFAR-10 train dataset. Test accuracy of non-trained models, i.e. models with random weights, is shown in the \emph{Accuracy (random model)} column. Test accuracy of these same models after training is shown in the \emph{Accuracy (trained model)} column. The models are 18-layer deep ResNet networks, and neuron activations of 10000 test samples extracted from the penultimate layer with 512 neurons are used as views. The effectivness of \texttt{ExItNeRdoM} methodology analysis is performed on redescriptions obtained from pairs of these views.}
    \begin{tabular}{c|c|c}
    Model           & Accuracy (random model) & Accuracy (trained model) \\ \hline
    resnet-18-s-211 & 0.090           & 0.853            \\
    resnet-18-s-212 & 0.100           & 0.876            \\
    resnet-18-s-213 & 0.099           & 0.881            \\
    resnet-18-s-214 & 0.098           & 0.845            \\
    resnet-18-s-215 & 0.100           & 0.837            \\
    resnet-18-s-216 & 0.100           & 0.872            \\
    resnet-18-s-217 & 0.100           & 0.872            \\
    resnet-18-s-218 & 0.100           & 0.859            \\
    resnet-18-s-219 & 0.100           & 0.859            \\
    resnet-18-s-220 & 0.100           & 0.866            \\
    resnet-18-s-221 & 0.100           & 0.879            \\
    resnet-18-s-222 & 0.100           & 0.873            \\
    resnet-18-s-223 & 0.100           & 0.877            \\
    resnet-18-s-224 & 0.100           & 0.872            \\
    resnet-18-s-225 & 0.100           & 0.870            \\
    resnet-18-s-226 & 0.100           & 0.843            \\
    resnet-18-s-227 & 0.100           & 0.872            \\
    resnet-18-s-228 & 0.104           & 0.868            \\
    resnet-18-s-229 & 0.100           & 0.858            \\
    resnet-18-s-230 & 0.098           & 0.868           
    \end{tabular}
    \label{tab:rand-cifar}
    \end{table}

\section{Rule extraction experiments (Section 6.2.)}

    \begin{table}[!htb]
    \centering
    \scriptsize
    \caption{Test performance of the MLP models on the ADNI dataset used in the rule extraction experiment. This table provides test accuracy for each of the 5 trained models produced by $5$-fold cross-validation. The MLP architecture consists of 33 input features; two hidden layers with 64 and 24 neurons, respectively; and 5 output neurons.}
    \begin{tabular}{c|c}
    Model        & Accuracy \\ \hline
    mlp-fold-1 & 0.685    \\
    mlp-fold-2 & 0.634    \\
    mlp-fold-3 & 0.678    \\
    mlp-fold-4 & 0.678    \\
    mlp-fold-5 & 0.689   
    \end{tabular}
    \label{tab:rule-adni}
    \end{table}

    \begin{table}[!htb]
    \centering
    \scriptsize
    \caption{Test performance of the MLP models on the WDBC dataset used in the rule extraction experiment. This table provides test accuracy for each of the 5 trained models produced by $5$-fold cross-validation. The MLP consists of 30 input features; two hidden layers with 64 and 32 neurons, respectively; and 1 output neuron.}
    \begin{tabular}{c|c}
    Model        & Accuracy \\ \hline
    mlp-fold-1 & 0.86     \\
    mlp-fold-2 & 0.895    \\
    mlp-fold-3 & 0.904    \\
    mlp-fold-4 & 0.921    \\
    mlp-fold-5 & 0.938   
    \end{tabular}
    \label{tab:rule-wdbc}
    \end{table}

\section{Explaining and relating layers of DLMs experiments  (Section 6.3.)}

    \begin{table}[!htb]
    \centering
    \scriptsize
    \caption{Architectures and test performance of computer vision (CV) models on the MNIST dataset used in experiments. Representations of 10000 test samples are extracted from the penultimate layer of each model, i.e. the final hidden layer before output layer. The \emph{Test Accuracies} column demonstrate the performance of each model on the MNIST test set. In the \emph{Input Size} column, $[x,y,z]$ represents the shape of input data, where $x$ is number of channels, $y$ is width, and $z$ is height. In the \emph{Hidden Layers} column, ($h_j$) represent the $j^{th}$ hidden layer; and $[x,y,z]$ denotes the shape of data representation after each hidden layer, where $x$ is number of channels, $y$ is width, and $z$ is height.}
    \begin{tabular}{c|c|c|c|c}
    \textbf{Model} & \textbf{Test Accuracy} & \textbf{Input Size} & \textbf{Hidden Layers}                                                                                                       & \textbf{Outputs} \\ \hline
    MLP            & 0.9793                 & 784                 & 250 ($h_1$), 100 ($h_2$)                                                                                                           & 10              \\\hline
    CNN-1          & 0.9882                 & $[1,28,28]$           & $[16, 20, 20]$ ($h_1$), 256 ($h_2$)                                                                                                  & 10              \\\hline
    CNN-2          & 0.9902                 & $[1,28,28] $          & $[16, 20, 20]$ ($h_1$), $[32, 6, 6]$   ($h_2$), 256 ($h_3$)                                                                               & 10              \\\hline
    CNN-3          & 0.9928                 & $[1,28,28]$           & \begin{tabular}[c]{@{}c@{}}$[8, 24, 24]$ ($h_1$), $[16, 20, 20]$ ($h_2$),\\ $[64, 6, 6]$ ($h_3$), 256 ($h_4$)\end{tabular}                     & 10              \\\hline
    CNN-4          & 0.9924                 & $[1,28,28]$           & \begin{tabular}[c]{@{}c@{}}$[8, 24, 24]$ ($h_1$), $[16, 20, 20]$ ($h_2$), \\ $[16, 16, 16]$ ($h_3$), $[64, 6, 6]$ ($h_4$), 256 ($h_5$)\end{tabular} & 10             
    \end{tabular}

    \label{tab:mnist}
    \end{table}

    \begin{table}[!htb]
    \centering
    \scriptsize
    \caption{Test performance of ResNet-18 models trained on CIFAR-10 used in experiments. Each initialized with different random seeds (111-114). The models are 18 layers deep, and representations are taken from the penultimate layer with 512 neurons. The test set contains 10000 samples. The test accuracies reflect the variation in performance across different initializations.}
    \begin{tabular}{c|c}
    \textbf{Model}         & \textbf{Test Accuracy} \\ \hline
    ResNet-18   (Seed 111) & 0.872                  \\
    ResNet-18   (Seed 112) & 0.8768                 \\
    ResNet-18   (Seed 113) & 0.8788                 \\
    ResNet-18   (Seed 114) & 0.882                 
    \end{tabular}
    \label{tab:cifar}
    \end{table}

    \begin{table}[!htb]
    \centering
    \scriptsize
    \caption{Test performance of BERT models on AG-NEWS dataset used in experiments. Two pre-trained models were used, Bert-mini and Bert-base, fine-tuned on the AG-NEWS dataset. Representations are extracted from both the start-of-sequence (SOS) token and the end-of-sequence (EOS) token. The test set consists of 7600 samples.}
    \begin{tabular}{c|c}
    \textbf{Model}          & \textbf{Test Accuracy} \\ \hline
    Bert-mini               & 0.934                  \\
    Bert-base               & 0.941                  \\
    \end{tabular}
    \label{tab:bert}
    \end{table}

\section{The \texttt{ExitNeRdoM} settings file}

\begin{verbatim}
JavaPath = /usr/bin/java -> java path
ClusPath = /.../ExperimentsInterpretableDL/CLUSNewest.jar -> CLUS path
OutputFolder = /.../ExperimentsInterpretableDL/ -> output folder path
Input1 = /.../ExperimentsInterpretableDL/CIFAR10_resnet18s111_layer_5.arff
Input2 = /.../ExperimentsInterpretableDL/CIFAR10_resnet18s112_layer_5.arff -> views paths
Input3 = /.../ExperimentsInterpretableDL/CIFAR10_resnet18s113_layer_5.arff
Input4 = /.../ExperimentsInterpretableDL/CIFAR10_resnet18s114_layer_5.arff
preferenceFilePath = /.../ExperimentsInterpretableDL/preferences.txt -> preferences paths 
[JSImp PValImp AttDivImp ElemDivImp ECoverage] -> [0.2 0.15 0.15 0.15 0.15 0.2] used 
System = linux -> OS (Linux used)
clusteringMode = 2 -> use the number of bins computed using FD rule
legacy = false -> use the newest CLUS implementation
jsType = qnm -> Jaccard index type (no missing values, thus just regular Jaccard)
numRandomRestarts = 10 -> 10 different random initializations used for alternations
optimizationType = extraction -> redescriptions are extracted from larger candidate sets
numIterations = 1 -> use one alternation step per random restart
ATreeDepth = 8 -> use PCTs of depth 8
NumTarget = 100 -> divide rules into groups of 100 and transform to targets
numTrees = 1 -> use one main PCT model
W1SideTrees = classification
W2SideTrees = classification -> utilize classification trees
W3SideTrees = classification
W4SideTrees = classification
clusteringMemory = 10000 -> max memory in MB allowed to construct PCTs
numNewAttr = 1 -> rules considered if they differ by at least one attribute
redesSetSizeType = exact -> produce output redescription sets of predefined size
numRetRed = 300 -> return 300 redescriptions in each set
WorkingRSSize = 500 -> keep at least 500 candidates in memory
MaxRSSize = 1500 -> keep maximally 1500 candidates in memory
minimizeRules = yes
joiningProcedure = yes -> use conjunctive refinement
unguidedExpansion = yes -> allow unguided rule matching to increase candidate number
GeneratingModelType = PCT -> use PCT as a main model
numSupplementTrees = 2 -> use 2 supplementing trees
SupplementPredictiveTreeType = PCTSub -> use random subsets as ensemble type
allowSERed = no -> dissallow multiple redescriptions describing the same set of entities
MinSupport = 10 -> minimal required support set size
MaxSupport = 4000 -> maximal allowed support set size
minJS =0.3 -> minimal required Jaccard index
minAddRedJS = 0.1 -> minimal Jaccard required to use redescription for refinement
maxPval= 0.01 -> maximal p-value
allowLeftNeg = true
allowRightNeg = true
allowLeftDisj = true -> allow disjunctions and negations
allowRightDisj = true
ruleSizeNormalization = 20 -> use 20 attributes as a normalizing factor
OutputFileName = redescriptionsCIFARMW1_2_3_4Forest 

\end{verbatim}

\subsection{Parameters that varied between experiments}

Majority of parameters were fixed as in the aforementioned settings file, used on the CIFAR-10 dataset (the experiment is denoted ResNet). The main changes are: $200$ redescriptions are returned in each set on the MNIST$_{CNN_{1,2,3}}$ experiment, $300$ on the ResNet and $1000$ on all other experiments. The size of the resulting redescription sets was reduced for experiments performed on multi-view datasets to increase execution speed. WorkingRSSize and MaxRSSize were set to $10000$ and $30000$ on all experiments except the multi-view ResNet and MNIST$_{CNN_{1,2,3}}$, where it was set to $500$ and $1500$ again with the goal of increasing execution speed. Datasets used to create redescriptions, obtained from the sets used to test the DLM models, were reduced to $5000$ entities on the multi-view ResNet and the MNIST$_{CNN_{1,2,3}}$ experiments.
    
{\footnotesize\bibliography{mybibfile}}